%% file: NIPS_caps_a.tex
\documentclass{article}

\PassOptionsToPackage{numbers, compress}{natbib}
 \usepackage[preprint]{neurips_2026}

\usepackage[utf8]{inputenc} 
\usepackage[T1]{fontenc}    
\usepackage{hyperref}       
\usepackage{url}            
\usepackage{booktabs}       
\usepackage{amsfonts}       
\usepackage{nicefrac}       
\usepackage{microtype}      
\usepackage{xcolor}         
\usepackage{multirow}

\usepackage{enumitem}
\usepackage[utf8]{inputenc} 
\usepackage[T1]{fontenc}    
\usepackage{hyperref}       
\usepackage{url}            
\usepackage{booktabs}       
\usepackage{amsfonts}       
\usepackage{nicefrac}       
\usepackage{microtype}      
\usepackage{xcolor}         
\usepackage{xpatch}
\usepackage{amsmath,amssymb,mathtools}
\usepackage{graphicx}
\usepackage{amsfonts} 

\usepackage{amsthm}  
\usepackage{float}
\usepackage{rotating}
\usepackage{algorithm}
\usepackage{algpseudocode}

\newtheorem{proposition}{Proposition}
\usepackage{float}
\usepackage{amsthm,amsmath,amsfonts}

\newcommand{\best}[1]{\textbf{\textcolor{red}{#1}}}
\newcommand{\second}[1]{\textcolor{blue}{\underline{#1}}}
\newcommand{\mystar}{\textsuperscript{*}}

\usepackage{dblfloatfix}

\title{Supervision Recovery for Time Series 
	Anomaly Detection via Context-Anchored Pairing}

\author{%
	\textbf{Yifei Gao}$^{1}$ \quad
	\textbf{Tian Lan}$^{1}$ \quad
	\textbf{Yimeng Lu}$^{1}$ \quad
	\textbf{Xuming An}$^{1}$ \\
	\textbf{Meng Wang}$^{2}$ \quad
	\textbf{Wenjun He}$^{2}$ \quad
	\textbf{Yijie Li}$^{2}$ \quad
	\textbf{Chen Zhang}$^{1}$\thanks{Corresponding author.} \\
	{\normalfont $^{1}$Department of Industrial Engineering, Tsinghua University} \\
	{\normalfont $^{2}$Huawei} \\
	{\normalfont\texttt{gao-yf@mail.tsinghua.edu.cn}} \\
	{\normalfont\texttt{\{lant23,luym25,axm24\}@mails.tsinghua.edu.cn}} \\
	{\normalfont\texttt{\{wangmeng71,hewenjun8,liyijie5\}@huawei.com}} \\
	{\normalfont\texttt{zhangchen01@tsinghua.edu.cn}}
}

\begin{document}

\maketitle

\begin{abstract}
	Time series anomaly detection (TSAD) remains challenging not only
	because anomaly labels are scarce, but also because temporal anomalies
	are highly context-dependent. Existing methods often rely on
	unsupervised objectives or surrogate abnormal patterns, providing
	limited supervision for context-dependent normal--anomalous
	distinctions. We propose Context-Anchored Pair Supervision (CAPS),
	a supervision-recovery framework for TSAD. CAPS views ideal anomaly
	supervision as a matched comparison between normal and anomalous
	outcomes under the same temporal context, and seeks to recover such
	supervision without target-domain anomaly labels. Using simulated
	normal--anomalous pairs, CAPS learns structure and anomaly-semantic
	representations through reconstruction, background consistency, and
	within-pair counterfactual recombination. The resulting anomaly
	representations form a continuous semantic space with coarse modes
	and induce a sampleable multimodal prior. CAPS conditionally realizes
	sampled semantics as residual-form effects on target reference
	trajectories. The resulting context-anchored normal--anomalous
	counterparts provide temporal supervision for discriminative detector
	learning. Experiments on nine datasets show that CAPS achieves the
	strongest aggregate performance across all four evaluation metrics
	among the compared methods, while complementary ablations and transfer
	analyses support the roles of context anchoring, semantic
	disentanglement, and conditional realization.
\end{abstract}

\section{Introduction}

Time series anomaly detection (TSAD) aims to identify abnormal events
in sequential data, with broad applications in industrial monitoring
\citep{nizam2022real,chen2022deep}, system operations
\citep{audibert2020usad,guo2024logformer}, and diverse sensor-driven
systems \citep{su2019robust,xu2021anomaly,deng2021graph}.
Despite its importance, effective TSAD remains challenging not only
because anomaly labels are scarce, but also because temporal anomalies
are highly context-dependent \citep{mueller2025open,wang2025survey}.
The same local pattern may be benign in one temporal regime but
abnormal in another; for example, a sudden increase can be normal
during system start-up but anomalous during steady operation.
To address these challenges, existing TSAD methods often rely on
indirect or surrogate training signals, including reconstruction
\citep{shen2021time}, forecasting \citep{zhou2024kan}, density
estimation \citep{zhou2023detecting}, representation learning
\citep{park2026paano}, self-supervised or contrastive learning,
pseudo anomalies, synthetic labels, and anomaly injection
\citep{darban2025carla,obata2025robust}. Although effective in many
cases, these methods either infer anomaly evidence indirectly from
proxy objectives or obtain abnormal supervision from surrogate
patterns. The resulting training signal may not always align with
the normal--anomalous distinction relevant to the target temporal
context, particularly when surrogate abnormal patterns are
constructed without accounting for that context. This raises a key
question: \textbf{can we recover better supervision for TSAD without
	target-domain anomaly labels?}

Addressing this question first requires clarifying
\textbf{what such supervision should represent}. Inspired by a causal
perspective, we view an anomaly as the effect of an anomaly mechanism
acting on temporal evolution. Let \(X^{(0)}\) denote the normal
outcome of a sequence under a given temporal context and
\(X^{(\alpha)}\) the alternative outcome when anomaly mechanism
\(\alpha\) acts under the same context. Their observation-level
difference \(R^{(\alpha)}:=X^{(\alpha)}-X^{(0)}\) represents the
observable anomaly effect relative to the matched normal outcome.
This contrast does not specify a structural model of how the anomaly
is physically generated or assume that the underlying mechanism is
additive. The desired supervision is therefore a matched
normal--anomalous comparison under the same temporal context, such
that the discriminative contrast is concentrated on anomaly-induced
variation. Such paired outcomes, however, are not jointly observable
in real target-domain data. Simulated source data provide a practical route because controllable
normal--anomalous correspondences can be constructed by design,
without assuming that target anomalies follow the same predefined
categories. Yet directly using simulated anomalous observations as
target positives may mix anomaly effects with source--target
background differences, since these observations retain
source-specific temporal structure. Moreover, the same anomaly
semantics may manifest differently across temporal contexts, as
their effects depend on the reference dynamics. We therefore need
to extract transferable anomaly semantics from simulation and
re-realize them under target temporal contexts. This raises the
central methodological question: \textbf{how can transferable anomaly
semantics be learned from simulated correspondences and converted
into matched target-domain supervision?}

Building on this perspective, we propose Context-Anchored Pair
Supervision (CAPS), a framework for recovering target-domain
supervision in TSAD. CAPS first learns structure and anomaly-semantic
representations from matched simulated pairs through reconstruction,
background consistency, and counterfactual recombination.
Specifically, the recombination objective requires an anomalous
observation to remain reconstructible after replacing its structure
code with that of its matched normal reference, while retaining its
anomaly-semantic code and support. This encourages anomaly
information to remain usable with structure extracted from a normal
observation. The learned anomaly representations form a shared
continuous semantic space and induce a sampleable multimodal prior.
A context-conditioned residual generator then realizes sampled
semantics as anomaly effects on target reference trajectories.
Pairing each generated counterpart with its target reference yields
context-anchored temporal supervision for discriminative detector
learning. Our main contributions are summarized as follows:

\begin{itemize}
	\item We formulate TSAD as a supervision-recovery problem,
	seeking matched normal--anomalous supervision under a shared
	temporal context without target-domain anomaly labels.
	
	\item We propose CAPS, which learns anomaly-semantic
	representations from simulated pairs through reconstruction,
	background consistency, and counterfactual recombination,
	organizes them into a continuous semantic prior, and
	conditionally realizes sampled semantics on target references
	to construct context-anchored matched supervision.
	
	\item We conduct extensive experiments on nine datasets, where
	CAPS achieves the strongest aggregate performance among the
	compared methods. Complementary ablations and transfer analyses
	support the components of supervision recovery, while
	theoretical results characterize the roles of background
	mismatch, semantic coverage, and contextual realization in
	supervision transfer.
\end{itemize}

\section{Related Work}

\paragraph{Time Series Anomaly Detection.}
In label-scarce TSAD, methods typically learn from unlabeled or normal-dominated data. A common paradigm is to model regular temporal behavior and identify anomalies as deviations from learned normality, including reconstruction-based \citep{malhotra2016lstm,kim2024model}, forecasting-based \citep{munir2018deepant}, density-estimation \citep{zong2018deep}, and one-class methods \citep{scholkopf1999support}. Self-supervised, contrastive, and representation-learning approaches further improve temporal representations through auxiliary objectives or transformed views \citep{sanchez2025review,yue2022ts2vec}. While effective, these methods primarily derive anomaly evidence indirectly from learned normality or surrogate objectives. To obtain more explicit anomaly-oriented supervision, recent studies exploit limited anomaly labels or contextual anomalies \citep{pang2019deep,ruff2019deep,carmona2021neural,liu2024arc}, construct pseudo anomalies or synthetic labels \citep{obata2025robust}, apply degradation-based or contrastive transformations \citep{jeong2023anomalybert,darban2025carla}, or introduce artificial anomalies into normal sequences \citep{shentu2024towards}. Such approaches reduce dependence on anomaly annotations, but their induced abnormal patterns are often governed by predefined perturbations or transformation rules, which may limit diversity and adaptability to temporal contexts and do not explicitly learn transferable anomaly semantics from matched normal--anomalous correspondences.

Beyond rule-based anomaly construction, learning-based generative approaches using GANs, diffusion models, and flow models \citep{li2019mad,yuan2024diffusion,hu2024flowts} provide flexible mechanisms for modeling time-series distributions and generating time-series samples. However, directly modeling target-domain anomalous distributions typically requires observed anomaly examples, anomaly-conditioned data, or externally specified abnormal conditions, which may be unavailable in the target-label-scarce setting considered here. Using anomalous data from simulated or external domains reduces dependence on target anomaly examples, but transferring such observations or distributions may entangle anomaly characteristics with source-specific temporal structure rather than isolate transferable knowledge. In visual anomaly detection, mature vision-language and text-to-image models provide rich semantic priors for synthesizing unseen defects from normal images \citep{sun2025unseen}. Constructing a comparable prior for temporal anomalies is challenging because their observable effects depend on temporal dynamics, operating states, and the surrounding reference context. Transferring anomaly knowledge therefore requires distinguishing reusable anomaly semantics from their context-dependent realization. This raises a more fundamental question for TSAD: what can provide an appropriate prior for temporal anomaly variation, and how can transferable anomaly knowledge be extracted from such a prior and adapted to target temporal contexts?

\section{TSAD as Target-Domain Supervision Recovery}
\label{sec:problem}

\noindent\textbf{Setting and available information.}
Let \(x=(x_1,\ldots,x_T)\in\mathbb{R}^T\) denote a time-series
window. We consider target-domain TSAD where the training set
\(\mathcal{D}^{\mathrm{tr}}_{\mathrm{tar}}
=\{x_i^{\mathrm{tar}}\}_{i=1}^{N_t}\)
is unlabeled and normal-dominated, and the goal is to learn a
detector \(D_\theta\) without target-domain anomaly labels.
We additionally assume access to simulated normal--anomalous data
with controllable correspondences, including matched normal and
anomalous outcomes, anomaly supports, and coarse simulated
anomaly-family labels. Such data provide structured information
about anomaly-induced variation, but constitute neither
target-domain anomaly labels nor target-domain anomalous
observations. The simulated family labels provide a coarse
organization of anomaly variation without prescribing exhaustive
anomaly categories in the target domain. Given our focus on
temporal anomaly behavior, we primarily consider anomaly effects
over time without explicitly modeling inter-variable dependencies.

\noindent\textbf{Desired supervision under temporal context.}
To characterize the supervision missing in the target domain,
let \(X_C^{(0)}\) denote the normal outcome under temporal context
\(C\), and \(X_C^{(\alpha)}\) the outcome when a conceptual anomaly
mechanism \(\alpha\in\mathcal{A}\) acts under the same context.
Their observation-level contrast
\(R_C^{(\alpha)}:=X_C^{(\alpha)}-X_C^{(0)}\)
describes the realized anomaly effect relative to the matched
normal outcome. Here, \(\alpha\) denotes the conceptual mechanism,
whereas \(R_C^{(\alpha)}\) is its observable manifestation under
context \(C\); hence, the same mechanism may manifest differently
across contexts, with
\(R_{C_1}^{(\alpha)}\neq R_{C_2}^{(\alpha)}\)
in general. This difference is an outcome-level representation of
the matched contrast: it neither assumes an additive underlying
mechanism nor requires causal identification of \(\alpha\).
The desired anomaly-oriented supervision is therefore the matched
comparison \((X_C^{(0)},X_C^{(\alpha)})\), whose shared temporal
context isolates anomaly-induced variation from unrelated
background differences.

\noindent\textbf{Why simulated supervision is not directly transferable.}
Although simulation provides such matched correspondences,
simulated anomalies are realized under simulated rather than
target temporal contexts. For a simulated pair
\((x^{\mathrm{sim}}_{n,j},x^{\mathrm{sim}}_{a,j})\)
and an unrelated target reference \(x_i^{\mathrm{tar}}\),
\(x^{\mathrm{sim}}_{a,j}-x_i^{\mathrm{tar}}
=(x^{\mathrm{sim}}_{a,j}-x^{\mathrm{sim}}_{n,j})
+(x^{\mathrm{sim}}_{n,j}-x_i^{\mathrm{tar}})\),
where the two terms correspond to the source-context anomaly
effect and source--target background mismatch. Directly using a
simulated anomaly as a target positive therefore confounds
anomaly-induced variation with reference mismatch. Moreover, even
\(x^{\mathrm{sim}}_{a,j}-x^{\mathrm{sim}}_{n,j}\)
is a realization tied to its simulated context and need not match
how the same anomaly knowledge manifests on a target trajectory.
Thus, the transferable object is neither the source anomalous
observation nor its realized effect itself. CAPS instead extracts
anomaly semantics that remain meaningful across contexts and
re-realizes them according to target temporal dynamics.

\noindent\textbf{Goal: target-domain supervision recovery.}
Given the simulated correspondences and
\(\mathcal{D}^{\mathrm{tr}}_{\mathrm{tar}}\), our goal is to
construct target-matched pairs
\(\widehat{\mathcal{P}}_{\mathrm{tar}}
=\{(x_i^{\mathrm{tar}},\widehat{x}_{i,a},m_i)\}\),
where \(\widehat{x}_{i,a}\) preserves the temporal background of
\(x_i^{\mathrm{tar}}\) while realizing learned anomaly semantics
on support \(m_i\). Under the normal-dominated assumption, target
windows serve as nominal reference anchors rather than being
assumed strictly anomaly-free. The reference and its constructed
counterpart provide temporal supervision \(\mathbf{0}\) and
\(m_i\), respectively, where \(m_i\) is known by construction
rather than observed from the target domain; the resulting
supervision is therefore temporal rather than merely window-level.
Accordingly, \emph{supervision recovery} means constructing
target-compatible normal--anomalous training supervision, not
recovering unknown ground-truth labels, exact anomalous outcomes,
or physical anomaly mechanisms. This formulation yields two
complementary requirements: extracting transferable anomaly
semantics from structured simulation and realizing them according
to target temporal contexts.

\section{Context-Anchored Pair Supervision}
\label{sec:overview}

\begin{figure*}[t]
	\centering
	\includegraphics[width=0.92\textwidth]{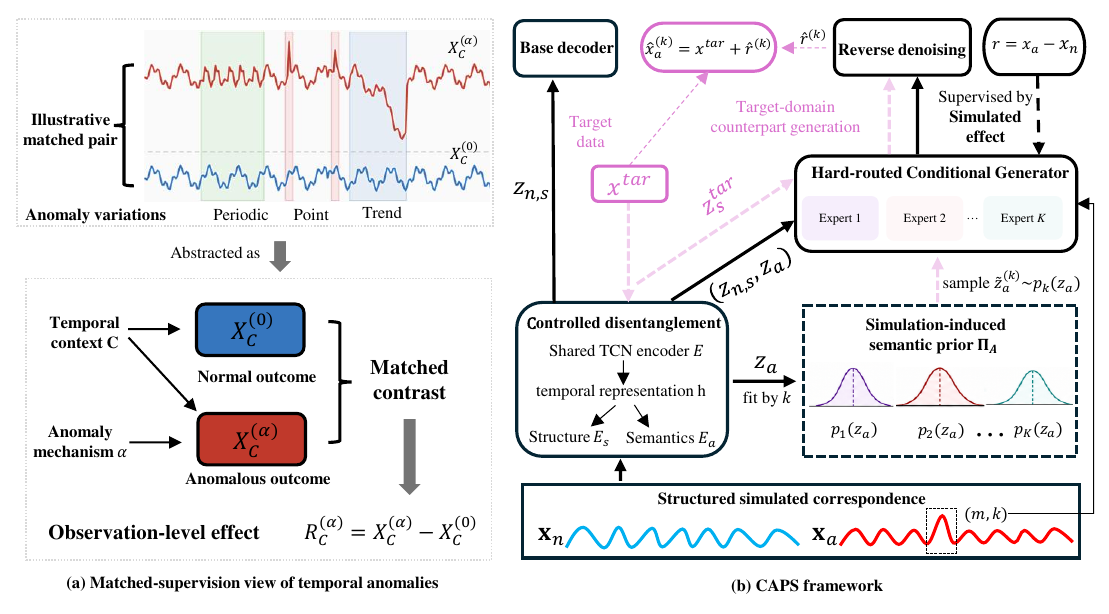}
	\caption{Overall illustration of CAPS. Panel (a) illustrates the
		matched-supervision view of temporal anomalies. Panel (b) presents the CAPS framework: black arrows denote learning
		from simulated correspondences, while purple arrows denote
		target-domain counterpart generation.}
	\label{fig:CAPS}
	\vspace{-0.05in}
\end{figure*}

Sec.~\ref{sec:problem} formulates TSAD as target-domain supervision
recovery, requiring transferable anomaly semantics and their realization
under target temporal contexts. CAPS addresses these requirements in a
unified framework. First, it learns structure and anomaly-semantic
representations from simulated normal--anomalous pairs through
reconstruction, background consistency, and counterfactual recombination.
In particular, replacing the structure code of an anomalous sample with
that of its matched normal reference encourages the anomaly-semantic
code to remain usable with the normal structure. The resulting semantic
representations induce a sampleable multimodal prior. Second, CAPS
organizes this prior into coarse semantic modes and employs a
hard-routed residual MoE to realize sampled semantics on target reference
trajectories, allowing experts to specialize in heterogeneous modes.
Given a target reference \(x_i^{\mathrm{tar}}\), the generator produces
a context-dependent anomaly effect \(\hat{r}_i^{(k)}\) and constructs
its matched counterpart as
\(\hat{x}_{i,a}^{(k)}=x_i^{\mathrm{tar}}+\hat{r}_i^{(k)}\).
Together with the sampled anomaly support, each reference--counterpart
pair provides recovered temporal supervision for discriminative detector
learning. CAPS therefore transfers anomaly knowledge through learned
semantic representations and target-conditioned realization.

\subsection{Structured Disentanglement for Transferable Anomaly Semantics}
\label{sec:disentangle}

\noindent\textbf{Learning from matched temporal contrasts.}
Sec.~\ref{sec:problem} distinguishes transferable anomaly semantics
from their context-dependent observation-level effects. CAPS learns
this factorization from simulated normal--anomalous pairs. Within each
pair, the temporal reference is shared, while the anomaly introduces
a controlled change. This correspondence provides a reference for
learning which temporal structure should be preserved and which
variation is associated with the anomaly. We exploit it through
complementary constraints on reconstruction, background consistency,
and recombination of the learned factors.

Following the simulated anomaly generation procedure
of~\citep{lan2025towards}, we construct correspondences
\((x_n,x_a,m,k)\), where \(x_n\) is the normal reference,
\(x_a\) its anomalous counterpart, \(m\in\{0,1\}^T\) denotes
anomaly support, and \(k\) indexes a coarse simulated anomaly family.
This procedure is used to construct the simulated source
correspondences; CAPS subsequently learns representations for semantic
transfer and target-conditioned realization. We learn two complementary
representations: \(z_s\) captures temporal structure to be preserved,
while \(z_a\in\mathcal{Z}_A\) captures anomaly information in a shared
continuous semantic space. The family index \(k\) provides a coarse
source-side organization of this space, allowing continuous variation
within each family without prescribing exhaustive target anomaly
categories. The observable realization of anomaly information is
jointly determined by its semantic representation and the temporal
structure on which it acts.

We adopt a shared TCN-based encoder \(E\) to capture multi-scale temporal
context through hierarchical receptive fields while retaining local
anomaly-related patterns. As shown in Fig.~\ref{fig:CAPS}(b), \(E\)
is followed by a structure head \(E_s\) and an anomaly-semantic head
\(E_a\). Auxiliary base and context-conditioned residual decoders
provide reconstruction constraints. Given a matched pair,
\(h_n=E(x_n)\) and \(h_a=E(x_a)\), from which we obtain
\(z_{n,s}=E_s(h_n)\), \(z_{a,s}=E_s(h_a)\),
\(z_{n,a}=E_a(h_n)\), and \(z_a=E_a(h_a)\).

\noindent\textbf{Reconstruction and background consistency.}
The base decoder reconstructs temporal structure, while the residual
decoder reconstructs the anomaly effect conditioned on both structure
and anomaly semantics. Let
\[
\hat{x}_n=\mathrm{base}(z_{n,s}),
\qquad
\hat{x}_a=
\mathrm{base}(z_{a,s})+\mathrm{res}(z_{a,s},z_a,m).
\]
The reconstruction loss is
\begin{equation}
	\mathcal{L}_{\mathrm{rec}}
	=
	\frac{1}{2}
	\left[
	\mathrm{MSE}(\hat{x}_n,x_n)
	+
	\mathrm{MSE}(\hat{x}_a,x_a)
	\right].
	\label{eq:lrec}
\end{equation}
Conditioning residual reconstruction on temporal structure allows the
observable effect to depend on the reference context.

To constrain the role of the structure branch, we define
\[
\mathrm{maskedL1}(u,v,q)
=
\frac{\sum_t q_t|u_t-v_t|}
{\sum_t q_t+\eta},
\]
where \(\eta>0\) is a small constant, and use
\begin{equation}
	\mathcal{L}_{\mathrm{base}}
	=
	\mathrm{MSE}(\mathrm{base}(z_{n,s}),x_n)
	+
	\mathrm{maskedL1}
	\big(\mathrm{base}(z_{a,s}),x_a,1-m\big).
	\label{eq:lbase}
\end{equation}
This objective reconstructs the normal reference and preserves the
observed background outside the anomaly support. Because the two
members of a pair share the same temporal reference, their structure
representations should remain consistent. We also suppress
anomaly-semantic activation for the normal member:
\begin{equation}
	\mathcal{L}_{\mathrm{dis}}
	=
	\|z_{n,s}-z_{a,s}\|_2^2
	+
	\|z_{n,a}\|_2^2.
	\label{eq:ldis}
\end{equation}

\noindent\textbf{Counterfactual recombination through structure swapping.}
To further constrain how the two representations are used, CAPS
requires the anomalous outcome to remain reconstructible after
swapping its structure code with that of the matched normal reference.
Specifically, we replace \(z_{a,s}\) with \(z_{n,s}\) while retaining
the anomaly-semantic code \(z_a\) and support \(m\):
\[
\mathrm{Dec}(z_{n,s},z_a,m)
=
\mathrm{base}(z_{n,s})
+
\mathrm{res}(z_{n,s},z_a,m).
\]
The matched anomalous observation \(x_a\) provides a reconstruction
target for this swapped combination, yielding
\begin{equation}
	\mathcal{L}_{\mathrm{cf}}
	=
	\mathrm{MSE}
	\big(
	\mathrm{Dec}(z_{n,s},z_a,m),
	x_a
	\big).
	\label{eq:lcf}
\end{equation}

This loss directly trains the decoder to combine structure extracted
from a normal observation with anomaly information extracted from its
matched anomalous counterpart. It complements
\(\mathcal{L}_{\mathrm{dis}}\): the latter encourages agreement between
the paired structure codes, whereas \(\mathcal{L}_{\mathrm{cf}}\)
requires the swapped factors to reconstruct the anomalous outcome.
The recombination constraint thus encourages \(z_a\) to carry anomaly
information that remains usable with the normal structure code, while
\(z_{n,s}\) supplies the context for its realization. This within-pair
training provides a basis for the subsequent generation stage, where
sampled anomaly semantics are combined with target-reference structure.

The complete representation-learning objective is
\[
\mathcal{L}_{\mathrm{disen}}
=
\mathcal{L}_{\mathrm{rec}}
+
\mathcal{L}_{\mathrm{base}}
+
\mathcal{L}_{\mathrm{dis}}
+
\mathcal{L}_{\mathrm{cf}}.
\]
Together, these objectives encourage accurate reconstruction,
background consistency, and recomposability of the learned factors.
The resulting anomaly-semantic representations form the basis of the
sampleable prior used for target-conditioned anomaly-effect generation.

\subsection{Context-Conditioned Anomaly-Effect Generation}
\label{sec:generation}

\noindent\textbf{Simulation-induced semantic prior and expert specialization.}
The paired representation learning in
Sec.~\ref{sec:disentangle} yields anomaly-semantic codes
\(z_a\) in a shared continuous space \(\mathcal{Z}_A\).
To make these semantics sampleable without target anomaly codes,
we fit a diagonal Gaussian
\(p_k(z_a)=\mathcal{N}(\mu_k,\Sigma_k)\)
for each coarse simulated anomaly family \(k\), yielding
\begin{equation}
	\Pi_A(z_a)
	=
	\sum_{k=1}^{K}\pi_k p_k(z_a),
	\qquad
	p_k(z_a)=\mathcal{N}(\mu_k,\Sigma_k).
	\label{eq:semantic_prior}
\end{equation}
We use uniform weights \(\pi_k=1/K\). The \(K\) components provide
a coarse multimodal approximation within the same semantic space
rather than defining separate or exhaustive target anomaly categories.
Continuous variation within each mode is represented by \(z_a\),
with samples \(\tilde{z}_a^{(k)}\sim p_k(z_a)\) serving as
mode-specific semantic conditions. CAPS employs a hard-routed
residual MoE: during simulated training, family index \(k\) routes
each pair to expert \(U_k\), while \(z_a\) conditions residual
generation. A lightweight router, trained with the same family
supervision, maps \((z_{n,s},z_a)\) from simulated correspondences
to normalized expert-association scores for optional target-guided
mode selection. Standard CAPS instead uses all modes
(Sec.~\ref{sec:pairing}).

\noindent\textbf{Context-conditioned residual realization.}
The semantic prior captures transferable anomaly variation but does
not determine its context-specific manifestation. For a simulated
pair \((x_n,x_a)\), we define its realized effect as \(r=x_a-x_n\),
an observation-level contrast rather than an additive assumption
on the underlying anomaly mechanism. CAPS applies conditional
diffusion directly in this observation-level residual space.
At diffusion step \(\tau\), we form
\(r_\tau=\sqrt{\bar{\alpha}_\tau}r
+\sqrt{1-\bar{\alpha}_\tau}\epsilon\),
where \(\epsilon\sim\mathcal{N}(0,I)\) and
\(\bar{\alpha}_\tau\) denotes the cumulative signal-retention
coefficient of the forward diffusion process.
Expert \(U_k\) predicts
\begin{equation}
	\hat{\epsilon}^{(k)}
	=
	U_k(r_\tau,\tau,z_{n,s},z_a,m).
	\label{eq:uk}
\end{equation}
Here, \(z_a\), \(z_{n,s}\), and \(m\) specify anomaly semantics,
reference context, and intended support, respectively. Thus, \(z_a\)
specifies what is transferred, while \(z_{n,s}\) determines how it
is realized under the reference dynamics. For Eq.~\ref{eq:lgen},
the corresponding clean-residual estimate is computed as
\(
\hat r^{(k)}
=
(r_\tau-\sqrt{1-\bar{\alpha}_\tau}\hat{\epsilon}^{(k)})
/
\sqrt{\bar{\alpha}_\tau}.
\)

For a target reference \(x_i^{\mathrm{tar}}\), CAPS extracts
\(z_{i,s}^{\mathrm{tar}}=E_s(E(x_i^{\mathrm{tar}}))\).
For each selected mode \(k\), it samples
\(\tilde{z}_a^{(k)}\sim p_k(z_a)\) and mask guidance \(m\)
from the corresponding empirical mask pool.
Conditioned on \((z_{i,s}^{\mathrm{tar}},\tilde{z}_a^{(k)},m)\),
expert \(U_k\) produces \(\hat{r}_i^{(k)}\) through reverse
denoising rather than transferring a source residual.
Here, \(m\) specifies the intended support rather than an observed
target label and subsequently provides point-wise supervision for
the generated counterpart. Conditioning on \(z_{i,s}^{\mathrm{tar}}\)
allows the same semantics to realize differently across target
temporal contexts.

\noindent\textbf{Residual localization objective.}
Simulated pairs provide both \(r=x_a-x_n\) and support \(m\).
With \(\mathrm{maskedMSE}\) defined analogously to
\(\mathrm{maskedL1}\), we optimize
\begin{equation}
	\mathcal{L}_{\mathrm{gen}}
	=
	\underbrace{
		\mathrm{maskedMSE}\bigl(\hat{\epsilon}^{(k)},\epsilon,m\bigr)
	}_{\mathcal{L}_{\mathrm{main}}}
	+
	\underbrace{
		\mathrm{maskedMSE}\bigl(\hat{r}^{(k)},0,1-m\bigr)
	}_{\mathcal{L}_{\mathrm{out}}}.
	\label{eq:lgen}
\end{equation}
Only expert \(U_k\) is optimized for a family-\(k\) sample.
The first term supervises noise prediction within anomaly support,
while the second suppresses the clean-residual estimate outside it,
keeping unrelated regions anchored to the reference.
Thus, \(\mathcal{L}_{\mathrm{gen}}\) learns localized,
context-dependent realizations of sampled anomaly semantics.

\subsection{Target-Reference Pairing for Detector Supervision}
\label{sec:pairing}

\noindent\textbf{Constructing target-matched counterparts.}
Given the target-conditioned effect \(\hat{r}_i^{(k)}\), CAPS constructs the anomalous counterpart
\begin{equation}
	\hat{x}_{i,a}^{(k)}
	=
	x_i^{\mathrm{tar}}
	+
	\hat{r}_i^{(k)}.
	\label{eq:xobs_counterpart}
\end{equation}
The resulting pair shares the same target temporal background, with its controlled difference determined by the generated effect. Without target anomaly-mode guidance, standard CAPS instantiates all \(K\) semantic modes for each reference, exposing the detector to diverse effects represented by the learned semantic prior.

\noindent\textbf{Targeted mode selection with limited anomaly guidance.}
When a small amount of anomalous target data is available, CAPS(Diagnosis) uses it only to narrow the modes used for counterpart generation. For each known anomalous sequence \(x\in\mathcal{D}_{\mathrm{known}}\), the router input is constructed as
\(
(E_s(E(x)),E_a(E(x)))
\),
without requiring a matched normal counterpart, and the router produces normalized mode-association scores \(q_k(x)\). We aggregate these scores as
\(
\bar q_k
=
|\mathcal{D}_{\mathrm{known}}|^{-1}
\sum_{x\in\mathcal{D}_{\mathrm{known}}}q_k(x)
\)
and rank the modes accordingly. Let \(k_{(1)},\ldots,k_{(K)}\) denote this ordering. We select the smallest subset
\begin{equation}
	\mathcal{K}_{\rho}
	=
	\{k_{(1)},\ldots,k_{(J_\rho)}\},
	\qquad
	J_\rho
	=
	\min\left\{
	J:
	\sum_{j=1}^{J}\bar q_{k_{(j)}}
	\ge \rho
	\right\},
	\label{eq:diagnosis_modes}
\end{equation}
where \(\rho\in(0,1]\) is a cumulative-association threshold. Accordingly, \(\mathcal{K}_i=\{1,\ldots,K\}\) for standard CAPS and \(\mathcal{K}_i=\mathcal{K}_{\rho}\) for CAPS(Diagnosis). The known anomalies are used only for mode selection, not detector supervision, while cumulative selection avoids forcing the observed target anomaly variation into a single coarse mode.

\noindent\textbf{Recovered temporal supervision.}
With \(m_i^{(k)}\) denoting the support sampled during generation, the recovered target pair set is
\[
\widehat{\mathcal{P}}_{\mathrm{tar}}
=
\{(x_i^{\mathrm{tar}},\hat{x}_{i,a}^{(k)},m_i^{(k)}):
x_i^{\mathrm{tar}}\in\mathcal{D}^{\mathrm{tr}}_{\mathrm{tar}},
k\in\mathcal{K}_i\}.
\]
Because \(m_i^{(k)}\) is known by construction, CAPS provides temporal rather than window-level supervision: \(x_i^{\mathrm{tar}}\) is paired with label sequence \(\mathbf{0}\), while \(\hat{x}_{i,a}^{(k)}\) is paired with \(m_i^{(k)}\). We train an independent lightweight TCN detector \(D_\theta:\mathbb{R}^{T}\rightarrow[0,1]^{T}\),
\begin{equation}
	D_\theta(x)
	=
	\sigma\!\left(h_\theta(E_D(x))\right),
	\label{eq:detector}
\end{equation}
using point-wise binary cross-entropy over these constructed labels. They constitute target-compatible supervision by construction rather than recovered ground-truth annotations. We keep the detector lightweight so that performance gains primarily reflect the recovered supervision; the complete procedure is provided in Algorithm~\ref{alg:caps} of Appendix~\ref{algorithm}.

\subsection{Analysis of Supervision Transfer}
\label{sec:analysis}

\noindent\textbf{Reference matching.}
For a simulated pair, define its observation-level effect as
\(r^s:=x_a^s-x_n^s\). Relative to an unrelated target reference \(x\),
\begin{equation}
	x_a^s-x
	=
	\underbrace{r^s}_{\text{source-context effect}}
	+
	\underbrace{(x_n^s-x)}_{\text{background mismatch}}.
	\label{eq:bg_mismatch}
\end{equation}
This identity does not assume an additive anomaly mechanism.
CAPS constructs \(\hat{x}_a=x+\hat r\), removing the cross-reference
term from the matched contrast. Whether this provides useful
supervision still depends on which effects are covered and how
they are realized under the target context.

\noindent\textbf{Contextual realization.}
Consider an idealized effect \(R=g(C,Z)\), where \(C\) denotes
temporal context and \(Z\) denotes anomaly semantics and support
guidance. If \(\mathbb{E}\|R\|_2^2<\infty\), then
\begin{equation}
	\inf_h\mathbb{E}\|R-h(Z)\|_2^2
	=
	\mathbb{E}\!\left[
	\operatorname{tr}\operatorname{Cov}(R\mid Z)
	\right],
	\label{eq:context_effect_gap}
\end{equation}
with optimum \(h^*(Z)=\mathbb{E}[R\mid Z]\).
Thus, effect prediction without context incurs irreducible error
when the same semantics produce different effects across contexts.
The oracle predictor \(g(C,Z)\) attains zero approximation error.
This motivates target-conditioned realization, but does not
establish a detection advantage or a distributional lower bound
for stochastic residual transfer.

\noindent\textbf{Detector risk.}
Let \(\xi=(r,m)\in\Xi\) and
\(\ell_D(x,\xi)=\ell(D(x+r),m)\), where \(\ell\) is the temporal
classification loss. Given references \(x\sim P_X\), denote ideal
matched effect--support supervision by
\(P_{\mathrm{tar}}^\Xi(\cdot\mid x)\) and CAPS supervision by
\(Q_{\Pi_A}^\Xi(\cdot\mid x)\). Define
\begin{equation}
	\mathcal R_\nu(D)
	=
	\frac12\mathbb E_x
	\left[
	\ell(D(x),\mathbf 0)
	+
	\mathbb E_{\xi\sim\nu(\cdot\mid x)}
	\ell_D(x,\xi)
	\right].
	\label{eq:matched_training_risk}
\end{equation}
These distributions respectively induce
\(\mathcal R_{\mathrm{ideal}}\) and \(\mathcal R_{\mathrm{CAPS}}\),
sharing the same reference component.

Introduce \(Q_{\Pi_A}^{\Xi,*}\), the analytical distribution obtained
with CAPS's semantic prior and support sampling but ideal contextual
realization. Using a common metric \(d_\Xi\), define
\begin{equation}
	\begin{aligned}
		\Delta_{\mathrm{cov}}
		&:=\mathbb E_x W_1\!\left(
		P_{\mathrm{tar}}^\Xi(\cdot\mid x),
		Q_{\Pi_A}^{\Xi,*}(\cdot\mid x)
		\right),\\
		\Delta_{\mathrm{real}}
		&:=\mathbb E_x W_1\!\left(
		Q_{\Pi_A}^{\Xi,*}(\cdot\mid x),
		Q_{\Pi_A}^\Xi(\cdot\mid x)
		\right).
	\end{aligned}
	\label{eq:coverage_realization}
\end{equation}
The first captures coverage and sampling mismatch; the second contextual realization error.

\begin{proposition}[Supervision-transfer risk]
	\label{prop:transfer}
	Assume the risks are finite, the conditional distributions have
	integrable first moments, and \(\ell_D(x,\cdot)\) is
	\(L\)-Lipschitz under \(d_\Xi\), uniformly over
	\(D\in\mathcal H\) and almost every \(x\).
	Set \(\Delta=\Delta_{\mathrm{cov}}+\Delta_{\mathrm{real}}\).
	Then
	\begin{equation}
		\sup_{D\in\mathcal H}
		\left|
		\mathcal R_{\mathrm{CAPS}}(D)
		-
		\mathcal R_{\mathrm{ideal}}(D)
		\right|
		\le \frac L2\Delta.
		\label{eq:transfer_bound}
	\end{equation}
	If \(\hat D\in\mathcal H\) satisfies
	\(\mathcal R_{\mathrm{CAPS}}(\hat D)
	\le\inf_{D\in\mathcal H}\mathcal R_{\mathrm{CAPS}}(D)
	+\epsilon_{\mathrm{learn}}\), then
	\begin{equation}
		\mathcal R_{\mathrm{ideal}}(\hat D)
		-
		\inf_{D\in\mathcal H}\mathcal R_{\mathrm{ideal}}(D)
		\le L\Delta+\epsilon_{\mathrm{learn}}.
		\label{eq:detector_excess_risk}
	\end{equation}
\end{proposition}

Reference matching removes the cross-reference contrast term,
while coverage and realization govern the remaining supervision
discrepancy. Their reduction by CAPS requires empirical validation.
The bounds compare matched training risks with shared reference labels,
so they do not account for contamination from anomalous references.
Proofs, sufficient conditions for BCE, and the interpretation of
\(\epsilon_{\mathrm{learn}}\) are provided in
Appendix~\ref{appendix:theory}.

\section{Experiments}

We evaluate CAPS through benchmark comparisons, component ablations,
and analyses of supervision transfer. The main text summarizes overall
detection performance, selected ablations and controlled comparisons,
and qualitative observations of the learned representations and
cross-context semantic recombination.
Detailed per-dataset results are provided in
Appendix~\ref{appendix:detailed_results}.
Appendix~\ref{appendix:extended_experiments} provides detailed analyses
of these comparisons and additional investigations of semantic
transfer, coverage, realization mechanisms, robustness, and efficiency.

\begin{table}[t]
	\caption{Overall performance across nine TSAD benchmarks.
		Score and Rank denote the unweighted mean score and average
		dataset-wise rank, respectively. Avg.R averages the ranks across
		all four metrics. T1 and T2 count first-place and top-two results
		across the 36 dataset--metric pairs.
		Best and second-best aggregate results are bold and underlined.
		Detailed per-dataset results are reported in
		Appendix~\ref{appendix:detailed_results}.}
	\label{tab:overall_results}
	\centering
	\small
	\setlength{\tabcolsep}{3.0pt}
	\begin{tabular}{l*{4}{cc}ccc}
		\toprule
		\multirow{2}{*}{Method}
		& \multicolumn{2}{c}{Aff.-F}
		& \multicolumn{2}{c}{$\mathrm{F1}_{\mathrm{T}}$}
		& \multicolumn{2}{c}{Std.-F1}
		& \multicolumn{2}{c}{VUS-PR}
		& \multirow{2}{*}{Avg.R}
		& \multirow{2}{*}{T1}
		& \multirow{2}{*}{T2} \\
		\cmidrule(lr){2-3}
		\cmidrule(lr){4-5}
		\cmidrule(lr){6-7}
		\cmidrule(lr){8-9}
		& Score$\uparrow$ & Rank$\downarrow$
		& Score$\uparrow$ & Rank$\downarrow$
		& Score$\uparrow$ & Rank$\downarrow$
		& Score$\uparrow$ & Rank$\downarrow$
		& & & \\
		\midrule
		
		\multicolumn{12}{c}{\textit{Unsupervised TSAD baselines}} \\
		\midrule
		LOF
		& 72.26 & 8.94 & 28.19 & 7.44
		& 23.11 & 8.33 & 23.19 & 8.06
		& 8.19 & 0 & 2 \\
		IForest
		& 47.86 & 10.67 & 14.15 & 10.94
		& 15.65 & 9.72 & 18.72 & 8.67
		& 10.00 & 0 & 0 \\
		OmniAnomaly
		& 73.48 & 7.78 & 28.35 & 5.56
		& 23.82 & 6.67 & 25.04 & 6.28
		& 6.57 & 1 & 6 \\
		TranAD
		& 73.39 & 8.67 & 21.46 & 7.67
		& 19.00 & 8.44 & 22.54 & 7.28
		& 8.01 & 0 & 0 \\
		USAD
		& 67.84 & 8.89 & 26.64 & 5.28
		& 25.23 & 4.44 & 25.92 & 6.22
		& 6.21 & 3 & 5 \\
		AnomTrans.
		& 69.10 & 9.00 & 7.60 & 11.56
		& 11.76 & 11.11 & 12.53 & 11.78
		& 10.86 & 0 & 0 \\
		TimesNet
		& 77.90 & 6.22 & 16.77 & 10.72
		& 15.99 & 10.33 & 20.05 & 9.72
		& 9.25 & 1 & 1 \\
		FITS
		& 77.02 & 6.22 & 15.64 & 10.94
		& 15.53 & 9.67 & 18.68 & 10.22
		& 9.26 & 0 & 1 \\
		
		\midrule
		\multicolumn{12}{c}{\textit{TCN-based references}} \\
		\midrule
		TCN-AE
		& 78.19 & 6.00 & 30.63 & 5.22
		& 25.10 & 5.78 & 23.96 & 5.89
		& 5.72 & 0 & 1 \\
		TCN-Injection
		& \underline{80.30} & \underline{3.89}
		& \underline{36.66} & \underline{4.00}
		& \underline{32.36} & \underline{4.00}
		& \underline{33.87} & \underline{3.11}
		& \underline{3.75} & \underline{5} & \underline{16} \\
		TCN-Supervised
		& 78.72 & 6.00 & 31.34 & 4.33
		& 27.50 & 4.67 & 30.72 & 4.44
		& 4.86 & 0 & 3 \\
		TCN-Forecast
		& 77.07 & 7.44 & 29.02 & 5.89
		& 24.43 & 6.61 & 23.05 & 8.00
		& 6.99 & 0 & 2 \\
		
		\midrule
		\textbf{CAPS}
		& \textbf{85.33} & \textbf{1.28}
		& \textbf{46.53} & \textbf{1.44}
		& \textbf{42.91} & \textbf{1.22}
		& \textbf{46.82} & \textbf{1.33}
		& \textbf{1.32} & \textbf{26} & \textbf{35} \\
		\bottomrule
	\end{tabular}
\end{table}

\subsection{Experimental Setup}

\noindent\textbf{Datasets and evaluation.}
We evaluate CAPS on nine public TSAD datasets spanning diverse
domains and anomaly characteristics: IOPS, MGAB, NAB, Power, SED,
UCR, NEK, TODS, and YAHOO.
Performance is reported using Affiliation-F,
$\mathrm{F1}_{\mathrm{T}}$, Standard-F1, and VUS-PR, covering
complementary event-level, point-wise, and range-aware
evaluation perspectives.

\noindent\textbf{Baselines.}
We compare CAPS with classical methods
LOF~\citep{breunig2000lof} and IForest~\citep{liu2008isolation},
and deep models OmniAnomaly~\citep{su2019robust},
USAD~\citep{audibert2020usad}, TranAD~\citep{tuli2022tranad},
Anomaly Transformer~\citep{xu2021anomaly},
TimesNet~\citep{wu2022timesnet}, and FITS~\citep{xu2024fits}.
TCN-based references include TCN-AE, TCN-Forecast,
TCN-Injection~\citep{shentu2024towards}, and TCN-Supervised.
Appendix~\ref{appendix:detailed_results} additionally reports
CAPS(Diagnosis), which uses limited known anomalies only for expert
selection, without using them as detector-training labels.
Baseline descriptions and experimental protocols are provided in
Appendix~\ref{appendix:exp_setup}.

\noindent\textbf{Implementation details.}
Unless otherwise specified, CAPS uses the TCN-based encoder--decoder
and independent lightweight TCN detector described in
Sec.~\ref{sec:overview}, with sequence length 256 and
structure/anomaly latent dimensions of 128/48.
We use $K=3$ coarse semantic modes organized by simulated point,
periodic, and trend anomaly families.
Alternative mode constructions and different choices of $K$
are examined in Appendix~\ref{appendix:extended_experiments}.
Implementation details are provided in
Appendix~\ref{appendix:exp_setup}.

\noindent\textbf{Ablations and Controlled Comparisons.}
\label{sec:ablations}
Table~\ref{tab:caps_ablation} summarizes four comparisons using the
same TCN detector and evaluation protocol:
\emph{(1) Shuffled residuals} applies each generated residual to a
different target reference, shuffling its anomaly support together
with it while preserving the generator and number of generated
counterparts;
\emph{(2) Without residual parameterization} replaces residual-form
construction with direct anomalous-sequence generation while retaining
background-preservation supervision outside the intended anomaly
support;
\emph{(3) Direct source-residual injection} adds the observation-level
differences of the same simulated normal--anomalous pairs directly
to target references, jointly bypassing semantic learning and
target-conditioned realization;
\emph{(4) Without counterfactual loss} removes
$\mathcal{L}_{\mathrm{cf}}$ in Eq.~\ref{eq:lcf} while retaining
$\mathcal{L}_{\mathrm{rec}}$, $\mathcal{L}_{\mathrm{base}}$, and
$\mathcal{L}_{\mathrm{dis}}$, examining the contribution of
within-pair counterfactual recomposition.
Detailed analyses of these comparisons are provided in
Appendix~\ref{appendix:extended_experiments}.

\begin{table}[t]
	\centering
	\caption{Ablations and controlled comparisons across nine datasets.
		Each entry is the unweighted average of the corresponding
		dataset-level scores. Higher is better; the best results are bolded.
		Detailed analyses are provided in
		Appendix~\ref{appendix:extended_experiments}.}
	\label{tab:caps_ablation}
	\small
	\begin{tabular*}{\linewidth}
		{@{\extracolsep{\fill}}lcccc@{}}
		\toprule
		Variant
		& Affiliation-F
		& $\mathrm{F1}_{\mathrm{T}}$
		& Standard-F1
		& VUS-PR \\
		\midrule
		Shuffled residuals
		& 79.27 & 41.21 & 37.78 & 41.30 \\
		w/o residual parameterization
		& 82.76 & 40.96 & 38.35 & 42.27 \\
		Direct source-residual injection
		& 81.58 & 43.16 & 39.96 & 42.58 \\
		w/o counterfactual loss
		& 83.65 & 42.76 & 39.11 & 43.11 \\
		\midrule
		\textbf{CAPS}
		& \textbf{85.33}
		& \textbf{46.53}
		& \textbf{42.91}
		& \textbf{46.82} \\
		\bottomrule
	\end{tabular*}
\end{table}

\begin{figure}[t]
	\centering
	\includegraphics[width=\linewidth]{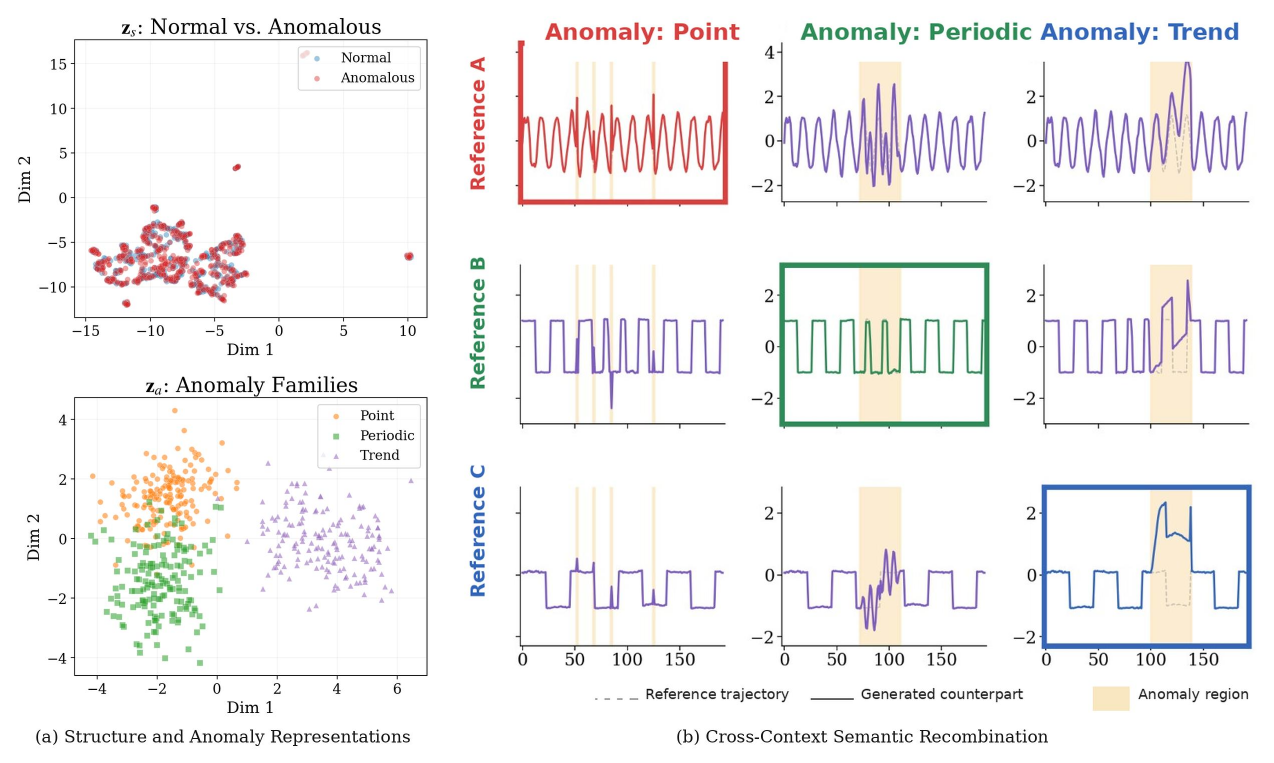}
	\caption{Learned representations and cross-context semantic recombination.
		(a) Projections of $z_s$ and $z_a$.
		(b) Each column reuses a fixed anomaly-semantic code across different
		references; colored diagonal cells indicate the original pairings.}
	\label{fig:cross_context_main}
\end{figure}

\subsection{Experimental Results}
\label{sec:results}

\noindent\textbf{Overall detection performance.}
Table~\ref{tab:overall_results} shows that CAPS achieves the strongest
aggregate performance across all four evaluation metrics among the
compared methods.
Its gains over conventional unsupervised methods support the value
of explicitly learning normal--anomalous distinctions through
recovered supervision.
The controlled TCN comparisons further examine this benefit while
keeping the detector backbone consistent.
In particular, the improvement over TCN-Injection highlights the
importance of how synthetic supervision is constructed.
By pairing each generated counterpart with its target reference,
CAPS provides a shared temporal background against which the detector
can learn anomaly-induced differences.
CAPS also outperforms the label-supervised TCN reference in aggregate
under the evaluated protocol.
Although this reference does not represent an upper bound on supervised
detection performance, the comparison illustrates the practical value
of recovering useful training supervision from simulated anomaly
knowledge when target anomaly labels are unavailable.

\noindent\textbf{Contributions of the supervision-recovery components.}
Table~\ref{tab:caps_ablation} shows that disrupting context--effect
matching, removing residual parameterization, or omitting the
counterfactual loss reduces aggregate detection performance.
These comparisons support matching generated effects to their
reference contexts, explicitly retaining the reference during
counterpart construction, and using within-pair recomposition
to learn representations for transfer.
The improvement over direct source-residual injection further supports
the combined value of semantic learning and target-conditioned
realization under the same simulated anomaly source.
Fig.~\ref{fig:cross_context_main} provides complementary qualitative
observations.
Panel~(a) shows substantial overlap between normal and anomalous
structure representations $z_s$, together with organization by
anomaly family in the anomaly-semantic representations $z_a$.
Panel~(b) illustrates how a fixed anomaly-semantic code can be
recombined with different reference structures, producing counterparts
that preserve the reference background while exhibiting
context-dependent anomaly effects.
These examples illustrate the intended distinction between transferable
anomaly semantics and their realization under a particular temporal
context. 

\section{Conclusion}
\label{sec:conclusion}

We proposed CAPS, a supervision-recovery framework for time series
anomaly detection without target-domain anomaly labels.
CAPS learns transferable anomaly semantics from simulated
normal--anomalous pairs, using within-pair counterfactual recomposition
to encourage their separation from temporal structure.
It conditionally realizes these semantics as residual effects on
target references, constructing context-anchored pairs for
discriminative detector learning.
Experiments on nine benchmarks and controlled comparisons support
the effectiveness of this approach and the contributions of its
core components.
Performance depends on how well the simulated semantic prior covers
target-relevant anomaly variation, motivating richer and adaptive
priors.
The current framework focuses on temporal anomaly effects;
explicit modeling of cross-variable dependencies remains an
important direction for future work.

\begin{ack}
	This work was supported by NSFC Grant 72271138, the Tsinghua
	University Initiative Scientific Research Program 20243080039,
	and the Tsinghua--Huawei Collaboration Project 20252001894.
	We also acknowledge computing support from the Tsinghua
	High-Performance Computing Center and the KaiTuo 1000
	supercomputer system.
\end{ack}

\bibliographystyle{unsrtnat}  
\small
\bibliography{caps}
\normalsize

\clearpage
\appendix

\section{Theoretical Details and Proofs}
\label{appendix:theory}

This section provides additional details for the theoretical analysis in the main text. We first formalize the limitation of predicting anomaly effects without temporal context, then prove Proposition~\ref{prop:transfer}, and finally clarify the regularity conditions, learning-error term, and scope of the resulting bound.

\subsection{Contextual Realization Gap}
\label{appendix:context_gap}

We first elaborate on why transferable anomaly semantics alone are generally insufficient to determine their observable effects. Let \(C\) denote temporal context, \(Z\) denote transferable anomaly information including semantic identity and support, and \(R\) denote the corresponding observation-level anomaly effect. Consider predictors \(h(Z)\) that predict the effect from transferable anomaly information without observing the temporal context.

Assuming finite second moments, the conditional mean
\[
h^{*}(Z)=\mathbb{E}[R\mid Z]
\]
minimizes the mean squared prediction error. By the conditional-variance decomposition,
\begin{align}
	\mathbb{E}\!\left[\|R-h(Z)\|_2^2\right]
	&=
	\mathbb{E}\!\left[
	\|R-\mathbb{E}[R\mid Z]\|_2^2
	\right]
	\nonumber\\
	&\quad+
	\mathbb{E}\!\left[
	\|\mathbb{E}[R\mid Z]-h(Z)\|_2^2
	\right].
\end{align}
Therefore,
\begin{equation}
	\inf_h
	\mathbb{E}\!\left[\|R-h(Z)\|_2^2\right]
	=
	\mathbb{E}\!\left[
	\operatorname{tr}\operatorname{Cov}(R\mid Z)
	\right].
	\label{eq:context_free_gap}
\end{equation}

Eq.~(\ref{eq:context_free_gap}) characterizes the minimum squared error
of predicting the corresponding effect from \(Z\) alone. In particular, when the same transferable anomaly information admits different conditional mean effects under different temporal contexts, context-independent prediction cannot fully recover the corresponding realized effect.

Allowing the predictor to additionally observe \(C\), the optimal predictor becomes
\[
f^{*}(C,Z)=\mathbb{E}[R\mid C,Z],
\]
with minimum error
\[
\inf_f
\mathbb{E}\!\left[
\|R-f(C,Z)\|_2^2
\right]
=
\mathbb{E}\!\left[
\operatorname{tr}\operatorname{Cov}(R\mid C,Z)
\right].
\]
The reduction in optimal squared error is
\begin{align}
	&
	\inf_h
	\mathbb{E}\!\left[
	\|R-h(Z)\|_2^2
	\right]
	-
	\inf_f
	\mathbb{E}\!\left[
	\|R-f(C,Z)\|_2^2
	\right]
	\nonumber\\
	&\qquad=
	\mathbb{E}\!\left[
	\left\|
	\mathbb{E}[R\mid C,Z]
	-
	\mathbb{E}[R\mid Z]
	\right\|_2^2
	\right]
	\ge 0.
	\label{eq:context_gain}
\end{align}
Hence, contextual conditioning improves the optimal squared-error prediction whenever temporal context changes the conditional mean realization of the same transferable anomaly information.

In the deterministic idealization used for intuition in the main text, where
\[
R=g(C,Z),
\]
a predictor with access to both \(C\) and \(Z\) can recover the corresponding effect exactly:
\[
\mathbb{E}\!\left[
\|R-g(C,Z)\|_2^2
\right]=0.
\]
This motivates realizing sampled anomaly semantics conditionally on the target reference trajectory rather than treating semantic identity as a context-independent residual template.

This analysis concerns matched-effect prediction under squared loss. It does not establish a distributional lower bound for stochastic residual transfer or, by itself, imply an advantage in downstream anomaly detection.

\subsection{Proof of Proposition~\ref{prop:transfer}}
\label{appendix:proof_transfer}

Let \(P_X\) denote the target-reference distribution and let
\[
\xi=(r,m)\in\Xi
\]
represent an anomaly-effect instance consisting of an observation-level residual \(r\) and its temporal support \(m\). For a target reference \(x\), define
\[
\ell_D(x,\xi)
=
\ell\!\left(D(x+r),m\right),
\qquad D\in\mathcal H,
\]
where \(\mathcal H\) is the detector hypothesis class and \(\ell\) is the temporal detection loss.

Let \(P_{\mathrm{tar}}^{\Xi}(\cdot\mid x)\) denote the ideal target-domain conditional distribution of anomaly effects and supports under reference \(x\), and let
\(Q_{\Pi_A}^{\Xi}(\cdot\mid x)\) denote the conditional distribution induced by CAPS. We additionally introduce an analytical intermediate distribution
\(Q_{\Pi_A}^{\Xi,*}(\cdot\mid x)\), which uses the same simulation-induced semantic prior and support sampling as CAPS but realizes them through an ideal target-context realization mechanism. Define
\begin{align}
	\Delta_{\mathrm{cov}}
	&=
	\mathbb{E}_{x\sim P_X}
	\left[
	W_1\!\left(
	P_{\mathrm{tar}}^{\Xi}(\cdot\mid x),
	Q_{\Pi_A}^{\Xi,*}(\cdot\mid x)
	\right)
	\right],
	\label{eq:delta_cov_appendix}
	\\
	\Delta_{\mathrm{real}}
	&=
	\mathbb{E}_{x\sim P_X}
	\left[
	W_1\!\left(
	Q_{\Pi_A}^{\Xi,*}(\cdot\mid x),
	Q_{\Pi_A}^{\Xi}(\cdot\mid x)
	\right)
	\right].
	\label{eq:delta_real_appendix}
\end{align}
The first measures semantic-coverage and sampling mismatch relative to the ideal target effect distribution, while the second measures realization discrepancy relative to this ideal mechanism. This decomposition is analytical and depends on the chosen intermediate distribution \(Q_{\Pi_A}^{\Xi,*}\); it is not intended as an identifiable decomposition of the two error sources from observed data.

For any conditional effect distribution \(\nu(\cdot\mid x)\), define the matched training risk
\begin{equation}
	\mathcal R_{\nu}(D)
	=
	\frac{1}{2}
	\mathbb{E}_{x\sim P_X}
	\left[
	\ell(D(x),\mathbf 0)
	+
	\mathbb{E}_{\xi\sim\nu(\cdot\mid x)}
	\ell_D(x,\xi)
	\right].
	\label{eq:matched_risk_appendix}
\end{equation}
We write
\[
\mathcal R_{\mathrm{ideal}}
=
\mathcal R_{P_{\mathrm{tar}}^{\Xi}},
\qquad
\mathcal R_{\mathrm{CAPS}}
=
\mathcal R_{Q_{\Pi_A}^{\Xi}}.
\]

Assume that, for \(P_X\)-almost every \(x\), the mapping
\[
\xi\mapsto\ell_D(x,\xi)
\]
is \(L\)-Lipschitz on \((\Xi,d_{\Xi})\), uniformly over
\(D\in\mathcal H\). Since the nominal reference term
\(\ell(D(x),\mathbf 0)\) is identical in the two risks,
\begin{align}
	&
	\left|
	\mathcal R_{\mathrm{CAPS}}(D)
	-
	\mathcal R_{\mathrm{ideal}}(D)
	\right|
	\nonumber\\
	&\quad\le
	\frac{1}{2}
	\mathbb{E}_{x\sim P_X}
	\left|
	\mathbb{E}_{\xi\sim Q_{\Pi_A}^{\Xi}(\cdot\mid x)}
	\ell_D(x,\xi)
	-
	\mathbb{E}_{\xi\sim P_{\mathrm{tar}}^{\Xi}(\cdot\mid x)}
	\ell_D(x,\xi)
	\right|.
\end{align}
By the Kantorovich--Rubinstein inequality,
\begin{align}
	\left|
	\mathcal R_{\mathrm{CAPS}}(D)
	-
	\mathcal R_{\mathrm{ideal}}(D)
	\right|
	&\le
	\frac{L}{2}
	\mathbb{E}_{x\sim P_X}
	W_1\!\left(
	P_{\mathrm{tar}}^{\Xi}(\cdot\mid x),
	Q_{\Pi_A}^{\Xi}(\cdot\mid x)
	\right).
	\label{eq:kr_bound}
\end{align}

For \(P_X\)-almost every \(x\), the triangle inequality for \(W_1\) gives
\begin{align}
	&
	W_1\!\left(
	P_{\mathrm{tar}}^{\Xi}(\cdot\mid x),
	Q_{\Pi_A}^{\Xi}(\cdot\mid x)
	\right)
	\nonumber\\
	&\quad\le
	W_1\!\left(
	P_{\mathrm{tar}}^{\Xi}(\cdot\mid x),
	Q_{\Pi_A}^{\Xi,*}(\cdot\mid x)
	\right)
	\nonumber\\
	&\qquad+
	W_1\!\left(
	Q_{\Pi_A}^{\Xi,*}(\cdot\mid x),
	Q_{\Pi_A}^{\Xi}(\cdot\mid x)
	\right).
\end{align}
Taking expectation over \(x\sim P_X\) therefore yields
\[
\mathbb{E}_{x\sim P_X}
W_1\!\left(
P_{\mathrm{tar}}^{\Xi}(\cdot\mid x),
Q_{\Pi_A}^{\Xi}(\cdot\mid x)
\right)
\le
\Delta_{\mathrm{cov}}
+
\Delta_{\mathrm{real}}.
\]
Consequently,
\begin{equation}
	\sup_{D\in\mathcal H}
	\left|
	\mathcal R_{\mathrm{CAPS}}(D)
	-
	\mathcal R_{\mathrm{ideal}}(D)
	\right|
	\le
	\frac{L}{2}
	\left(
	\Delta_{\mathrm{cov}}
	+
	\Delta_{\mathrm{real}}
	\right).
	\label{eq:transfer_bound_appendix}
\end{equation}

Let
\[
\Delta
=
\Delta_{\mathrm{cov}}
+
\Delta_{\mathrm{real}},
\]
and suppose that the learned detector \(\hat D\in\mathcal H\) satisfies
\begin{equation}
	\mathcal R_{\mathrm{CAPS}}(\hat D)
	\le
	\inf_{D\in\mathcal H}
	\mathcal R_{\mathrm{CAPS}}(D)
	+
	\epsilon_{\mathrm{learn}}.
	\label{eq:learning_error}
\end{equation}
Using Eq.(~\ref{eq:transfer_bound_appendix}),
\begin{align}
	\mathcal R_{\mathrm{ideal}}(\hat D)
	&\le
	\mathcal R_{\mathrm{CAPS}}(\hat D)
	+
	\frac{L}{2}\Delta
	\nonumber\\
	&\le
	\inf_{D\in\mathcal H}
	\mathcal R_{\mathrm{CAPS}}(D)
	+
	\epsilon_{\mathrm{learn}}
	+
	\frac{L}{2}\Delta
	\nonumber\\
	&\le
	\inf_{D\in\mathcal H}
	\mathcal R_{\mathrm{ideal}}(D)
	+
	L\Delta
	+
	\epsilon_{\mathrm{learn}}.
\end{align}
Hence,
\begin{equation}
	\mathcal R_{\mathrm{ideal}}(\hat D)
	-
	\inf_{D\in\mathcal H}
	\mathcal R_{\mathrm{ideal}}(D)
	\le
	L
	\left(
	\Delta_{\mathrm{cov}}
	+
	\Delta_{\mathrm{real}}
	\right)
	+
	\epsilon_{\mathrm{learn}},
\end{equation}
which proves Proposition~\ref{prop:transfer}.

\subsection{Additional Assumptions and Learning Error}
\label{appendix:theory_conditions}

We finally clarify the regularity conditions and scope of the preceding analysis.

\paragraph{Wasserstein regularity.}
The conditional distributions
\(P_{\mathrm{tar}}^{\Xi}(\cdot\mid x)\),
\(Q_{\Pi_A}^{\Xi,*}(\cdot\mid x)\), and
\(Q_{\Pi_A}^{\Xi}(\cdot\mid x)\)
are assumed to be measurable probability kernels with finite first moments under \(d_{\Xi}\). We additionally assume that
\[
\Delta_{\mathrm{cov}}
+
\Delta_{\mathrm{real}}
<\infty
\]
and that the matched risks in Eq.~(\ref{eq:matched_risk_appendix}) are finite. These conditions are sufficient for the Wasserstein quantities and expectations used in the proof to be well defined.

\paragraph{A sufficient Lipschitz condition for binary cross-entropy.}
The proposition requires the composed loss
\[
\xi\mapsto\ell(D(x+r),m)
\]
to admit a common Lipschitz constant uniformly over
\(D\in\mathcal H\). Binary cross-entropy is not globally Lipschitz when predicted probabilities approach \(0\) or \(1\). A sufficient condition is that, uniformly over \(D\in\mathcal H\), detector outputs on all relevant inputs satisfy
\[
D(u)\in[\delta,1-\delta]^T,
\qquad
0<\delta<\frac{1}{2},
\]
and that there exists a common constant \(K_D<\infty\) such that
\begin{equation}
	\frac{1}{T}
	\|D(u)-D(v)\|_1
	\le
	K_D\|u-v\|_2
	\label{eq:detector_lipschitz}
\end{equation}
for all relevant inputs \(u\) and \(v\).

Consider the averaged binary cross-entropy
\[
\ell_{\mathrm{BCE}}(p,m)
=
-\frac{1}{T}
\sum_{t=1}^{T}
\left[
m_t\log p_t
+
(1-m_t)\log(1-p_t)
\right].
\]
On \(p_t\in[\delta,1-\delta]\), its variation with respect to prediction satisfies
\[
\left|
\frac{\partial\ell_{\mathrm{BCE}}}{\partial p_t}
\right|
\le
\frac{1}{T\delta},
\]
while changing the binary label at a fixed prediction is bounded by
\[
\frac{1}{T}
\left|
\log\frac{1-p_t}{p_t}
\right|
\le
\frac{1}{T}
\log\frac{1-\delta}{\delta}.
\]

Under the product metric
\begin{equation}
	d_{\Xi}\!\left(
	(r,m),(r',m')
	\right)
	=
	\|r-r'\|_2
	+
	\frac{1}{T}
	\|m-m'\|_1,
	\label{eq:xi_metric}
\end{equation}
Eq.~(\ref{eq:detector_lipschitz}) therefore implies
\begin{align}
	&
	\left|
	\ell_{\mathrm{BCE}}\!\left(
	D(x+r),m
	\right)
	-
	\ell_{\mathrm{BCE}}\!\left(
	D(x+r'),m'
	\right)
	\right|
	\nonumber\\
	&\quad\le
	\frac{K_D}{\delta}
	\|r-r'\|_2
	+
	\log\frac{1-\delta}{\delta}
	\frac{1}{T}\|m-m'\|_1
	\nonumber\\
	&\quad\le
	L\,
	d_{\Xi}\!\left(
	(r,m),(r',m')
	\right),
\end{align}
where one valid common constant is
\begin{equation}
	L
	=
	\max\left\{
	\frac{K_D}{\delta},
	\log\frac{1-\delta}{\delta}
	\right\}.
\end{equation}
These conditions are only sufficient conditions for the theoretical bound; they need not be interpreted as additional constraints imposed by the CAPS training procedure.

\paragraph{Interpretation of \(\epsilon_{\mathrm{learn}}\).}
The term \(\epsilon_{\mathrm{learn}}\) represents ordinary detector-learning error under the CAPS-induced population risk rather than semantic-transfer discrepancy. To make this interpretation explicit, let
\(\widehat{\mathcal R}_{\mathrm{CAPS}}\) denote the empirical risk on the recovered training set, and suppose that
\[
\widehat{\mathcal R}_{\mathrm{CAPS}}(\hat D)
\le
\inf_{D\in\mathcal H}
\widehat{\mathcal R}_{\mathrm{CAPS}}(D)
+
\epsilon_{\mathrm{opt}}.
\]
Then
\begin{align}
	&
	\mathcal R_{\mathrm{CAPS}}(\hat D)
	-
	\inf_{D\in\mathcal H}
	\mathcal R_{\mathrm{CAPS}}(D)
	\nonumber\\
	&\quad\le
	2
	\sup_{D\in\mathcal H}
	\left|
	\widehat{\mathcal R}_{\mathrm{CAPS}}(D)
	-
	\mathcal R_{\mathrm{CAPS}}(D)
	\right|
	+
	\epsilon_{\mathrm{opt}}.
	\label{eq:learning_decomposition}
\end{align}
Thus, the right-hand side of Eq.~(\ref{eq:learning_decomposition})
can be taken as \(\epsilon_{\mathrm{learn}}\), separating
finite-sample estimation error from optimization error.

This deterministic decomposition does not itself specify a finite-sample convergence rate. Establishing such a rate would additionally require assumptions accounting for temporal dependence and for dependence among generated counterparts constructed from shared references. Moreover, because Proposition~\ref{prop:transfer} compares detectors within the same hypothesis class \(\mathcal H\), approximation error relative to an unrestricted Bayes-optimal detector is not included in \(\epsilon_{\mathrm{learn}}\).

\paragraph{Reference contamination.}
The bound compares ideal and CAPS matched-training risks under the same target-reference distribution and the same nominal reference label \(\mathbf 0\). It therefore isolates discrepancy associated with semantic coverage, sampling, and contextual realization. If a nominal target reference already contains an unobserved anomaly, assigning it the reference label \(\mathbf 0\) may introduce an additional supervision bias. Such contamination is not represented by
\(\Delta_{\mathrm{cov}}\) or
\(\Delta_{\mathrm{real}}\) and is outside the scope of Proposition~\ref{prop:transfer}; its empirical effect is examined separately in Appendix~\ref{appendix:supervision_analysis}.

\section{Training and Implementation Details}
\label{appendix:method_details}

\subsection{Overall Training and Inference Procedure}
\label{algorithm}

Algorithm~\ref{alg:caps} summarizes the training, supervision-recovery,
and inference procedure of CAPS.
CAPS first learns structure and anomaly-semantic representations
from matched simulated normal--anomalous pairs using the objective
in Sec.~\ref{sec:disentangle}.
Within-pair counterfactual recomposition reconstructs each anomalous
sample from its anomaly-semantic code and its matched normal
reference's structure code.
The representation modules are then fixed, mode-wise semantic priors
are estimated, and residual experts are trained.
During target-domain supervision recovery, sampled anomaly semantics
are conditionally realized on target references to construct matched
reference--counterpart pairs.
The detector is then trained with point-wise supervision on these
pairs and directly applied at inference.
In the standard setting, all modes are used without target anomaly
labels. With Diagnosis, limited known anomalies are used only to
select associated modes for targeted counterpart generation.
\begin{algorithm}[H]
	\caption{Overall Training, Supervision Recovery, and Inference Procedure of CAPS}
	\label{alg:caps}
	\begin{algorithmic}[1]
		\Require Simulated correspondences
		$\mathcal{D}_{\mathrm{sim}}$ containing $(x_n,x_a,m,k)$,
		unlabeled target training data
		$\mathcal{D}^{\mathrm{tr}}_{\mathrm{tar}}$,
		optional known anomalies $\mathcal{D}_{\mathrm{known}}$,
		Diagnosis threshold $\rho=0.8$ if enabled
		\Ensure Semantic priors $\{p_k\}_{k=1}^{K}$,
		residual experts $\{U_k\}_{k=1}^{K}$,
		detector $D_\theta$
		
		\State \textbf{Paired representation learning}
		\For{each training batch sampled from $\mathcal{D}_{\mathrm{sim}}$}
		\State Encode simulated normal--anomalous pairs to obtain
		$z_{n,s},z_{a,s},z_{n,a},z_a$
		\State Compute the within-pair recomposition loss
		$\mathcal{L}_{\mathrm{cf}}
		=\mathrm{MSE}(\mathrm{Dec}(z_{n,s},z_a,m),x_a)$
		\State Update the encoder, representation heads, and auxiliary
		decoders using
		$\mathcal{L}_{\mathrm{disen}}
		=\mathcal{L}_{\mathrm{rec}}
		+\mathcal{L}_{\mathrm{base}}
		+\mathcal{L}_{\mathrm{dis}}
		+\mathcal{L}_{\mathrm{cf}}$
		\EndFor
		\State Freeze the trained encoder $E$ and representation heads
		$E_s,E_a$
		
		\State \textbf{Mode organization and semantic-prior estimation}
		\State Establish mode assignments using the selected
		knowledge-based or data-driven organization
		\For{each mode $k=1,\ldots,K$}
		\State Encode anomaly samples assigned to mode $k$ and fit
		the diagonal Gaussian
		$p_k(z_a)=\mathcal{N}(\mu_k,\Sigma_k)$
		\State Build the empirical mask pool $\mathcal{M}_k$
		from correspondences assigned to mode $k$
		\EndFor
		\State Form
		$\Pi_A(z_a)=\sum_{k=1}^{K}\pi_kp_k(z_a)$
		with $\pi_k=1/K$
		
		\State \textbf{Residual-expert training}
		\If{Diagnosis is enabled}
		\State Train the lightweight router on paired representations
		$(z_{n,s},z_a)$ using the established simulated mode assignments
		to produce normalized mode-association scores
		\EndIf
		\For{each simulated training batch}
		\State Compute
		$z_{n,s}=E_s(E(x_n))$ and
		$z_a=E_a(E(x_a))$
		\State Form the observation-level effect $r=x_a-x_n$
		\State Hard-route each sample according to its established
		mode assignment $k$
		\State Update the selected expert using
		$\mathcal{L}_{\mathrm{gen}}$
		conditioned on $(z_{n,s},z_a,m)$
		\EndFor
		
		\State \textbf{Mode selection for target generation}
		\If{Diagnosis is enabled and $\mathcal{D}_{\mathrm{known}}\neq\emptyset$}
		\State Evaluate the router on
		$(E_s(E(x)),E_a(E(x)))$ for each
		$x\in\mathcal{D}_{\mathrm{known}}$
		to obtain $q_k(x)$
		\State Compute
		$\bar q_k\leftarrow
		\frac{1}{|\mathcal{D}_{\mathrm{known}}|}
		\sum_{x\in\mathcal{D}_{\mathrm{known}}}q_k(x)$
		for $k=1,\ldots,K$
		\State Sort modes such that
		$\bar q_{k_{(1)}}\ge\cdots\ge\bar q_{k_{(K)}}$
		\State Set
		$J_\rho\leftarrow
		\min\left\{
		J:\sum_{j=1}^{J}\bar q_{k_{(j)}}\ge\rho
		\right\}$
		\State Set
		$\mathcal{K}\leftarrow
		\{k_{(1)},\ldots,k_{(J_\rho)}\}$
		\Else
		\State Set $\mathcal{K}\leftarrow\{1,\ldots,K\}$
		\EndIf
		
		\State \textbf{Target-domain supervision recovery}
		\State Initialize an empty pair list
		$\widehat{\mathcal{P}}_{\mathrm{tar}}$
		\For{each target reference
			$x_i^{\mathrm{tar}}\in\mathcal{D}^{\mathrm{tr}}_{\mathrm{tar}}$}
		\State Extract
		$z_{i,s}^{\mathrm{tar}}
		=E_s(E(x_i^{\mathrm{tar}}))$
		\For{each $k\in\mathcal{K}$}
		\State Sample
		$\tilde z_a^{(k)}\sim p_k(z_a)$
		and $m\sim\mathcal{M}_k$
		\State Generate
		$\hat r_i^{(k)}$
		with $U_k$ conditioned on
		$(z_{i,s}^{\mathrm{tar}},\tilde z_a^{(k)},m)$
		\State Construct
		$\hat x_{i,a}^{(k)}
		=x_i^{\mathrm{tar}}+\hat r_i^{(k)}$
		\State Append
		$\big(
		(x_i^{\mathrm{tar}},\mathbf 0),
		(\hat x_{i,a}^{(k)},m)
		\big)$
		to $\widehat{\mathcal{P}}_{\mathrm{tar}}$
		\EndFor
		\EndFor
		
		\State \textbf{Detector training}
		\State Train
		$D_\theta:\mathbb{R}^{T}\rightarrow[0,1]^{T}$
		on $\widehat{\mathcal{P}}_{\mathrm{tar}}$
		using averaged point-wise BCE with equal weight on the
		reference and generated branches
		
        \State \textbf{Test-time scoring:} Output $D_\theta(x)$ for each test sequence $x$.
	\end{algorithmic}
\end{algorithm}

\subsection{Paired Representation Learning and Counterfactual Recomposition}
\label{appendix:paired_learning}

Representation learning uses simulated correspondences
\((x_n,x_a,m,k)\), where \(x_n\) is the normal reference,
\(x_a\) is its matched anomalous counterpart, \(m\) specifies the
anomaly support, and \(k\) denotes a coarse simulated anomaly family.
Each pair shares a temporal reference while exposing an
anomaly-induced change. The family label \(k\) provides the default
organization for subsequent prior estimation and expert training,
while the representation-learning objective operates on the
matched observations and their support.

As described in Sec.~\ref{sec:disentangle}, the shared encoder and
two representation heads produce \(z_{n,s}\), \(z_{a,s}\),
\(z_{n,a}\), and \(z_a\).
The reconstruction and background objectives constrain the
representations to explain the observations and preserve their
shared temporal background. The disentanglement objective
\(\mathcal L_{\mathrm{dis}}\) encourages agreement between
\(z_{n,s}\) and \(z_{a,s}\), while suppressing anomaly-semantic
activation \(z_{n,a}\) for the normal reference.

Within-pair counterfactual recomposition further constrains how
the learned factors are used. The ordinary anomalous reconstruction
uses the structure code \(z_{a,s}\) extracted from \(x_a\).
For recomposition, we replace this code with \(z_{n,s}\) from its
matched normal reference, while retaining \(z_a\) and \(m\):
\[
\mathrm{Dec}(z_{n,s},z_a,m)
=
\mathrm{base}(z_{n,s})
+
\mathrm{res}(z_{n,s},z_a,m).
\]
The reconstruction target remains the same matched anomalous
observation \(x_a\), giving
\[
\mathcal L_{\mathrm{cf}}
=
\mathrm{MSE}
\bigl(
\mathrm{Dec}(z_{n,s},z_a,m),x_a
\bigr).
\]
Both representations originate from the same simulated pair.
This constraint encourages anomaly information to remain usable
with structure extracted from a normal observation.
The residual decoder is conditioned on both factors, allowing
the reconstructed effect to depend on the reference context.

The complete representation-learning objective is
\[
\mathcal L_{\mathrm{disen}}
=
\mathcal L_{\mathrm{rec}}
+
\mathcal L_{\mathrm{base}}
+
\mathcal L_{\mathrm{dis}}
+
\mathcal L_{\mathrm{cf}}.
\]
All four terms are evaluated on matched-pair minibatches, and the
encoder, representation heads, and auxiliary decoders are optimized
jointly. After representation learning, the encoder and heads are
fixed for semantic-prior estimation and residual-expert training.

\subsection{Semantic Prior, Mode Organization, and Diagnosis}
\label{appendix:mode_details}

After paired representation learning, the encoder and representation
heads are fixed, yielding a shared continuous anomaly-semantic
space \(\mathcal Z_A\). The mode index \(k\) provides a coarse
organization of heterogeneous simulated anomaly semantics within
this common space; it does not define separate latent spaces or
an exhaustive taxonomy of target-domain anomalies.
Continuous semantic variation is retained within each mode through
\(z_a\), while mode organization provides a coarse specialization
structure for prior estimation and context-conditioned realization.

\paragraph{Mode organization.}
We consider two strategies for organizing the learned
anomaly-semantic space. The default strategy is
\emph{knowledge-based}. Based on the temporal characteristics of
the simulated anomaly operations, we organize them into three
coarse modes,
\[
K=3:
\qquad
\text{trend},\quad
\text{periodic},\quad
\text{point}.
\]
Here, \(K=3\) follows from this source-side organization rather
than being selected through target-domain supervision or
latent-space clustering. Each mode still contains continuous
variation in its operation-specific configuration parameters.

As an alternative, we consider a fully \emph{data-driven}
organization that does not use the predefined
trend--periodic--point assignments. After representation learning,
the frozen anomaly-semantic codes \(z_a\) are organized directly
in \(\mathcal Z_A\). We evaluate
\[
K\in\{1,3,5,7\},
\]
where \(K=1\) treats the semantic space as a single mode and
\(K>1\) uses \(K\)-means clustering to obtain the corresponding
mode assignments. These assignments replace the simulator-defined
mode labels throughout subsequent prior estimation and expert
specialization. For the clustering-based variants, the same
assignments also define the mode-specific mask pools and, when
applicable, the supervision labels for the Diagnosis router.

The two strategies therefore examine different sources of semantic
organization: the knowledge-based construction introduces a coarse
inductive bias derived from temporal anomaly structure, whereas
the data-driven construction derives the organization directly
from the learned semantic representations.
As evaluated in Appendix~\ref{appendix:coverage_analysis}, the
knowledge-based three-mode organization achieves the strongest
overall performance and is therefore adopted as the default setting.

Changing \(K\) in the data-driven variants changes not only the
specialization granularity but, under our default generation
protocol, also the total number of residual experts and the number
of generated counterparts per target reference. Comparisons across
different values of \(K\) should therefore be interpreted as
end-to-end mode-organization configurations rather than as an
isolated measure of specialization granularity.
In contrast, the knowledge-based and data-driven configurations
with \(K=3\) use the same number of modes and provide a more direct
comparison of the two organization strategies.

\paragraph{Mode-wise semantic prior.}
For a given mode organization, let
\[
\mathcal Z_k
=
\left\{
z_a^{(j)}: k_j=k
\right\}
\]
denote the anomaly-semantic codes assigned to mode \(k\).
We estimate a diagonal Gaussian
\[
p_k(z_a)
=
\mathcal N(\mu_k,\Sigma_k),
\]
where
\[
\mu_k
=
\frac{1}{|\mathcal Z_k|}
\sum_{z\in\mathcal Z_k}z,
\]
and \(\Sigma_k\) is the diagonal empirical covariance of the
corresponding semantic codes. The resulting simulation-induced
semantic prior is
\[
\Pi_A(z_a)
=
\sum_{k=1}^{K}\pi_kp_k(z_a),
\qquad
\pi_k=\frac{1}{K}.
\]
The Gaussian components provide a coarse multimodal approximation
within the same continuous semantic space rather than defining
independent anomaly spaces. Under the standard generation
procedure, each target reference receives one generated counterpart
from each mode, consistent with the uniform mixture weights.

\paragraph{Expert specialization.}
During simulated-domain expert training, each sample is hard-routed
according to its established mode assignment \(k\).
Under the knowledge-based organization, \(k\) is determined by the
simulated anomaly family; under the data-driven organization,
it is determined by the semantic-space clustering.
The corresponding residual expert \(U_k\) is optimized only with
samples assigned to that mode.
Such coarse specialization is intended to reduce interference
among heterogeneous semantic regions during context-conditioned
realization.

The lightweight router used for Diagnosis neither determines the
mode organization nor serves as the training gate for the residual
experts. Mode assignments are established before expert training,
whereas the router is used only to associate limited known target
anomalies with the already constructed modes.

\paragraph{Diagnosis with limited known anomalies.}
The standard CAPS setting requires no target-domain anomaly labels
and uses all available modes,
\[
\mathcal K=\{1,\ldots,K\}.
\]
When limited known target anomalies are available, CAPS(Diagnosis)
uses a lightweight router to estimate their associations with
the simulation-derived modes. During simulated training, the router
takes \((z_{n,s},z_a)\) from matched correspondences and is
supervised by their established mode assignments.
For a known target anomaly \(x\), its input is
\((E_s(E(x)),E_a(E(x)))\), so no matched normal counterpart is
required. It outputs normalized mode-association scores
\[
q_k(x)\ge0,
\qquad
\sum_{k=1}^{K}q_k(x)=1.
\]
Given a small known-anomaly set \(\mathcal D_{\mathrm{known}}\),
we compute
\[
\bar q_k
=
\frac{1}{|\mathcal D_{\mathrm{known}}|}
\sum_{x\in\mathcal D_{\mathrm{known}}}
q_k(x).
\]
Let
\[
\bar q_{k_{(1)}}\ge
\bar q_{k_{(2)}}\ge
\cdots\ge
\bar q_{k_{(K)}}
\]
denote the sorted associations.
We use a cumulative-association threshold
\[
\rho=0.8
\]
and define
\[
J_\rho
=
\min
\left\{
J:
\sum_{j=1}^{J}
\bar q_{k_{(j)}}\ge\rho
\right\},
\]
with the retained mode set
\[
\mathcal K_\rho
=
\left\{
k_{(1)},\ldots,k_{(J_\rho)}
\right\}.
\]
The selected modes are used equally for counterpart generation,
corresponding to
\[
\Pi_{A,\mathcal K_\rho}(z_a)
=
\frac{1}{|\mathcal K_\rho|}
\sum_{k\in\mathcal K_\rho}
p_k(z_a).
\]

The known anomalies are used only for mode association and targeted
counterpart generation rather than as detector-training labels.
The router is unnecessary in the standard CAPS setting and is not
used by the final detector at inference.

\subsection{Architecture and Optimization Details}
\label{appendix:implementation}

Table~\ref{tab:caps_hparams} summarizes the main architectural
and optimization configurations used in CAPS.
Unless otherwise specified, all reported results use an input
length of 256 and the knowledge-based three-mode organization
described in Appendix~\ref{appendix:mode_details}.
The representation and generation modules are trained in two stages.
We first optimize the paired representation-learning objective
and then freeze the encoder and representation heads.
The fixed representations are subsequently used to estimate the
semantic priors and train the hard-routed residual experts.
The detector is trained independently on the recovered
target-domain reference--counterpart pairs.

\paragraph{Two-stage representation and generation training.}
In the first stage, the shared encoder \(E\), structure head \(E_s\),
anomaly-semantic head \(E_a\), and auxiliary decoders are jointly
optimized on minibatches of matched simulated normal--anomalous
pairs using
\[
\mathcal L_{\mathrm{disen}}
=
\mathcal L_{\mathrm{rec}}
+
\mathcal L_{\mathrm{base}}
+
\mathcal L_{\mathrm{dis}}
+
\mathcal L_{\mathrm{cf}}.
\]
We use Adam with a learning rate of \(10^{-3}\), a batch size
of 32, and train the representation modules for 60 epochs.
After this stage, \(E\), \(E_s\), and \(E_a\) are frozen,
thereby fixing the learned structure and anomaly-semantic spaces.

In the second stage, the frozen representation modules are used
to extract \(z_{n,s}\) and \(z_a\) from the simulated
correspondences. The mode-wise diagonal Gaussian priors
\(\{p_k\}_{k=1}^{K}\) are estimated from these fixed semantic
codes, and the residual experts are optimized using
\(\mathcal L_{\mathrm{gen}}\).
Each simulated sample is hard-routed according to its established
mode assignment.
The residual-generation stage is trained for 100 epochs using
Adam with the same learning rate and batch size.
We use a diffusion horizon of 200 steps and run the full
200-step reverse process for counterpart generation.
The lightweight router is trained only when CAPS(Diagnosis)
is evaluated and does not participate in standard expert routing.
The base and residual decoders in Table~\ref{tab:caps_hparams}
are auxiliary reconstruction modules used during representation
learning.
The diffusion experts operate directly in observation-level
residual space; the clean-residual estimate defined in
Sec.~\ref{sec:generation} is used to evaluate
\(\mathcal L_{\mathrm{out}}\) during training, whereas
counterpart generation uses the full reverse diffusion process.

\begin{table}[H]
	\caption{Main architectural and optimization configurations of CAPS.}
	\label{tab:caps_hparams}
	\centering
	\small
	\setlength{\tabcolsep}{5pt}
	\renewcommand{\arraystretch}{1.05}
	\begin{tabular}{ll}
		\toprule
		\textbf{Component} & \textbf{Setting} \\
		\midrule
		Input shape
		& $[B,1,256]$ \\
		Default mode organization
		& $K=3$ (trend / periodic / point) \\
		Latent dimensions
		& $\dim(z_s)=128$, $\dim(z_a)=48$ \\
		Encoder
		& hidden dims $[64,128,256]$, dropout $0.1$ \\
		Multi-scale extractor
		& 4 branches; kernels $(3,5,7,9)$; dilations $(1,2,3,4)$ \\
		Residual experts
		& $K$ hard-routed experts ($K=3$ by default) \\
		Diagnosis router
		& hidden dim $128$, $\rho=0.8$ (optional) \\
		Location embedding
		& 64 \\
		Diffusion process
		& 200 training timesteps; 200 reverse sampling steps \\
		Diffusion U-Net
		& channels $[64,128,256,256]$ \\
		Base decoder width
		& 256 \\
		Residual decoder width
		& 192 \\
		Detector
		& independent TCN with point-wise sigmoid output \\
		\midrule
		Optimizer
		& Adam \\
		Learning rate
		& $1\times10^{-3}$ \\
		Batch size
		& 32 \\
		Representation-learning epochs
		& 60 \\
		Residual-generation epochs
		& 100 \\
		Detector-training epochs
		& 20 \\
		\bottomrule
	\end{tabular}
\end{table}

After residual-generator training, the representation and generation
modules remain fixed while target-domain counterparts are constructed.
The detector \(D_\theta\) is then optimized independently for
20 epochs using Adam with the same learning rate and batch size.
Training uses averaged point-wise binary cross-entropy on the
recovered pair list, with equal weight assigned to the reference
and generated branches.
This balances the two branches of each recovered pair but does
not imply balanced positive and negative labels across individual
time points.
No representation, generator, or semantic-prior parameters are
updated during detector training.

\paragraph{Reproducibility.}
To facilitate reproducibility, we provide an anonymized repository
containing the CAPS implementation, training and evaluation scripts,
configuration files for the main experiments, and instructions
for preparing the benchmark datasets:
\url{https://github.com/gyf-0623/caps}.
The configuration files specify the remaining implementation
details, including the diffusion noise schedule.
We additionally release the simulated normal--anomalous
correspondences, including their mode assignments, anomaly supports,
and reference identifiers, to support reproduction of paired
representation learning and subsequent supervision construction.
\section{Experimental Setup and Detailed Results}
\label{appendix:exp_setup}

\subsection{Benchmark Datasets}

We follow the TSB-AD benchmark~\citep{liu2024elephant} and conduct the
main evaluation on nine public TSAD datasets: IOPS, MGAB, NAB, NEK,
Power, SED, TODS, UCR, and YAHOO. We use the benchmark-defined training
and test partitions throughout. For standard CAPS, the training portion
serves as the unlabeled target-reference pool, with its anomaly labels
unused, while the test portion is reserved for evaluation. These datasets
span diverse application domains, including operations, sensors, web
services, traffic, weather, and energy, and exhibit heterogeneous anomaly
frequencies and temporal characteristics. Their basic statistics are
summarized in Table~\ref{tab:main_datasets}, where AR denotes the anomaly
ratio.

\begin{table}[t]
	\caption{Benchmark datasets used in the main experiments.}
	\label{tab:main_datasets}
	\centering
	\small
	\setlength{\tabcolsep}{5pt}
	\renewcommand{\arraystretch}{1.05}
	\begin{tabular}{lcccc}
		\toprule
		Name & Domain & \#TS & Avg. Length & AR (\%) \\
		\midrule
		UCR    & Misc.      & 228 & 67818.7 & 0.6 \\
		NAB    & Web        & 28  & 5099.7  & 10.6 \\
		YAHOO  & Web        & 259 & 1560.2  & 0.6 \\
		IOPS   & Operations & 17  & 72792.3 & 1.3 \\
		MGAB   & Sensor     & 9   & 97777.8 & 0.2 \\
		SED    & Energy     & 3   & 23332.3 & 4.1 \\
		TODS   & Traffic    & 15  & 5000.0  & 6.3 \\
		NEK    & Weather    & 9   & 1073.0  & 8.0 \\
		Power  & Power Grid & 1   & 35040.0 & 8.5 \\
		\bottomrule
	\end{tabular}
\end{table}

\subsection{Baseline Methods}

We compare CAPS with representative TSAD methods and controlled TCN-based training variants.

\paragraph{Conventional TSAD baselines.}
The external baselines include classical methods LOF~\citep{breunig2000lof} and IForest~\citep{liu2008isolation}, together with deep TSAD models OmniAnomaly~\citep{su2019robust}, USAD~\citep{audibert2020usad}, TranAD~\citep{tuli2022tranad}, Anomaly Transformer~\citep{xu2021anomaly}, TimesNet~\citep{wu2022timesnet}, and FITS~\citep{xu2024fits}. We follow the benchmark-recommended configurations whenever available. In particular, the input window is set to 10 for TranAD, 100 for USAD, and 100 for OmniAnomaly; LOF uses 50 neighbors, and IForest uses 200 estimators. The remaining benchmark baselines use their corresponding benchmark configurations.

\paragraph{Controlled TCN-based variants.}
To separate the effect of supervision construction from detector-backbone differences, we additionally construct four TCN-based variants using the same input length, preprocessing pipeline, and TCN detector backbone as CAPS. TCN-AE and TCN-Forecast use reconstruction and forecasting objectives, respectively. TCN-Injection employs perturbation-based synthetic anomaly training~\citep{shentu2024towards}.

TCN-Supervised serves as a same-backbone label-supervised reference. It is trained on the labeled target-domain training split using point-wise binary cross-entropy, with the original temporal labels preserved; test labels are used only for evaluation. Unlike CAPS, which constructs matched reference--counterpart pairs and explicitly balances the reference and generated branches during detector training, TCN-Supervised follows the empirical anomaly frequency of the labeled training data and may therefore be affected by the severe class imbalance common in TSAD. We consequently treat it as a controlled supervised reference rather than as an optimized upper bound on supervised performance. Additional imbalance-aware supervised variants are examined separately in Appendix~\ref{appendix:supervision_analysis}.

These controlled variants allow us to distinguish the contribution of recovered supervision from gains attributable solely to the TCN architecture or exposure to synthetic anomalous samples. In particular, TCN-Injection examines whether the construction of synthetic supervision matters beyond synthetic anomaly exposure, while TCN-Supervised provides a reference for learning from factual target-domain labels under the same detector architecture.

We also report CAPS(Diagnosis). In this variant, 10\% of anomalous windows
from the target test partition are treated as limited known anomalies and
are used only for mode association and targeted counterpart generation
through the Diagnosis mechanism described in Appendix~\ref{appendix:mode_details}.
They are not used as detector-training labels. Evaluation is performed on
the full test partition, including these known-anomaly windows; therefore,
CAPS(Diagnosis) is treated as a target-guided diagnostic extension rather
than part of the standard label-free setting.

\subsection{Evaluation Metrics}

We evaluate detection performance using Affiliation-F, \(\mathrm{F1}_{\mathrm{T}}\), Standard-F1, and VUS-PR. Following TSB-AD~\citep{liu2024elephant}, we use Affiliation-F, Standard-F1, and VUS-PR, and follow~\citep{sarfraz2024position} for \(\mathrm{F1}_{\mathrm{T}}\). Together, these metrics provide complementary point-wise, temporal, event-level, and range-aware perspectives on detection quality.

\begin{itemize}
	\item \textbf{Standard-F1.}
	Standard-F1 is the conventional point-wise F1 score computed from precision and recall over individual timestamps.
	
	\item \textbf{\(\mathrm{F1}_{\mathrm{T}}\).}
	\(\mathrm{F1}_{\mathrm{T}}\) evaluates temporal anomaly detection while accounting for the correspondence between predicted and ground-truth anomaly regions, making it less dependent on exact point-wise alignment than Standard-F1.
	
	\item \textbf{Affiliation-F.}
	Affiliation-F evaluates predicted and ground-truth anomaly events through affiliation-based temporal matching~\citep{huet2022local}, providing an event-oriented measure that is tolerant to moderate temporal misalignment.
	
	\item \textbf{VUS-PR.}
	VUS-PR is a threshold-independent and range-aware metric that integrates precision--recall performance over varying temporal tolerances~\citep{paparrizos2022volume}, thereby accounting for uncertainty in anomaly boundaries and detection delay.
\end{itemize}
\paragraph{Threshold selection and VUS configuration.}
Following the TSB-AD evaluation protocol, Standard-F1,
$\mathrm{F1}_{\mathrm{T}}$, and Affiliation-F are evaluated
using metric-specific oracle thresholds. For each test
sequence and metric, we search over 1,500 equally spaced
anomaly-score quantile levels and report the best value
obtained using the test labels. VUS-PR is computed directly
from continuous anomaly scores without selecting a single
operating threshold. Its window length is automatically
determined from the dominant period estimated using the
autocorrelation function of the test sequence.

\subsection{Implementation Settings}

Unless otherwise specified, CAPS uses an input length of 256 and the knowledge-based three-mode organization described in Appendix~\ref{appendix:mode_details}. The representation-learning, residual-generation, and detector-training stages are trained for 60, 100, and 20 epochs, respectively, using Adam with a learning rate of \(10^{-3}\) and batch size 32. The diffusion model uses 200 training timesteps and the full 200-step reverse process for counterpart generation. CAPS-specific architectural, semantic-prior, mode-organization, and Diagnosis settings are provided in Appendix~\ref{appendix:implementation}.

For all controlled TCN variants, the target-domain input construction, preprocessing procedure, and detector backbone are kept consistent with CAPS wherever applicable. Their differences therefore arise primarily from the training objective and the form of supervision provided to the detector. External benchmark methods retain their benchmark-specific model configurations rather than being forced into the CAPS input protocol.

\subsection{Detailed Per-Dataset Results}
\label{appendix:detailed_results}

Table~\ref{tab:detailed_results} reports the complete per-dataset results corresponding to the aggregate comparison in the main text. We report Affiliation-F, $\mathrm{F1}_{\mathrm{T}}$, Standard-F1, and VUS-PR on all nine benchmark datasets. CAPS and CAPS(Diagnosis) results are averaged over five runs. Table~\ref{tab:controlled_results_std} reports the mean and
standard deviation over five independent runs for CAPS and
the controlled TCN references. To keep the comparison with the standard label-free setting clear, CAPS(Diagnosis) is excluded when determining the best and second-best methods; we separately mark whether Diagnosis improves over standard CAPS for each dataset--metric pair.

The detailed results show that the aggregate advantage of CAPS is not driven by a single dataset or evaluation criterion. Standard CAPS remains particularly strong under both event-oriented and point-/range-aware metrics, while the controlled comparisons indicate that neither the shared TCN backbone nor synthetic anomaly exposure alone explains the observed gains. The results also illustrate substantial variation across datasets, reinforcing the value of reporting complementary metrics rather than relying on a single evaluation protocol.

CAPS(Diagnosis) improves over standard CAPS on many dataset--metric pairs, showing that limited known anomalies can provide useful information for targeted mode selection. The improvement is not universal, however. In particular, restricting generation to modes associated with a small known-anomaly subset can discard useful semantic coverage when that subset is not fully representative of the target anomaly distribution. We therefore treat Diagnosis as an optional target-guided extension rather than part of the standard label-free CAPS setting. 

\begin{table}[H]
	\caption{Performance on nine datasets under four metrics (higher is better). CAPS variants are averaged over five runs. Best and second-best results excluding \textbf{CAPS(Diagnosis)} are shown in bold and underline, respectively; a star marks the better result between \textbf{CAPS} and \textbf{CAPS(Diagnosis)} for each dataset--metric pair.}
	\label{tab:detailed_results}
	\centering
	\scriptsize
	\setlength{\tabcolsep}{2.2pt}
	\renewcommand{\arraystretch}{0.88}
	
	\resizebox{0.96\textwidth}{!}{%
		\begin{tabular}{ll*{9}{c}}
			\toprule
			Metric & Method & IOPS & MGAB & NAB & Power & SED & UCR & NEK & TODS & YAHOO \\
			\midrule
			
			\multirow{14}{*}{Affiliation-F}
			& LOF             & 81.06 & 68.44 & 75.75 & 66.76 & 63.85 & 73.53 & 84.74 & 60.58 & 75.63 \\
			& IForest         & 52.81 & 68.82 & 39.84 & 0.00 & 70.09 & 50.56 & 71.15 & 44.17 & 33.30 \\
			& OmniAnomaly     & 80.32 & 67.35 & \second{92.35} & 78.16 & 61.26 & 73.53 & 86.30 & 50.73 & 71.31 \\
			& TranAD          & 83.19 & 67.28 & 90.28 & 71.56 & 61.03 & 73.31 & 85.02 & 52.76 & 76.08 \\
			& USAD            & 71.08 & 67.81 & 91.54 & 76.48 & 55.60 & 76.00 & 71.13 & 47.90 & 53.05 \\
			& AnomTrans.      & 70.79 & 67.65 & 79.03 & 71.57 & 68.21 & 80.03 & 75.32 & 44.57 & 64.75 \\
			& TimesNet        & 85.04 & 66.93 & 89.24 & 69.08 & 66.98 & 80.55 & \best{89.79} & 70.36 & 83.13 \\
			& FITS            & 86.36 & 67.77 & 89.03 & 67.96 & 66.99 & 74.36 & \second{87.76} & 71.77 & 81.19 \\
			\cmidrule(lr){2-11}
			& TCN-AE          & 83.91 & 67.61 & 85.74 & 83.34 & 67.75 & 74.54 & 85.74 & \second{72.61} & 82.47 \\
			& TCN-Injection   & \second{87.58} & \second{69.13} & 88.89 & 85.04 & \second{70.94} & \second{84.32} & 81.52 & 72.05 & \second{83.22} \\
			& TCN-Supervised  & 84.35 & 67.59 & 89.62 & \second{85.96} & 68.47 & 72.40 & 85.24 & 71.75 & 83.11 \\
			& TCN-Forecast    & 76.05 & 67.27 & 84.06 & 84.80 & 67.68 & 72.81 & 85.83 & 72.48 & 82.64 \\
			\cmidrule(lr){2-11}
			& CAPS            & \best{89.24} & \best{69.22} & \best{93.25} & \best{85.98} & \best{74.09} & \best{88.41} & 86.30 & \best{86.85}\mystar & \best{94.63}\mystar \\
			\cmidrule(lr){2-11}
			& CAPS(Diagnosis) & 89.83\mystar & 69.61\mystar & 93.81\mystar & 86.35\mystar & 75.82\mystar & 89.72\mystar & 86.52\mystar & 81.32 & 94.57 \\
			\midrule
			
			\multirow{14}{*}{$\mathrm{F1}_{\mathrm{T}}$}
			& LOF             & 27.97 & 1.15 & 35.76 & 19.80 & 9.60 & 8.31 & 63.57 & \second{31.63} & 55.93 \\
			& IForest         & 7.64 & 0.84 & 21.44 & 0.00 & 9.54 & 6.36 & 65.56 & 11.06 & 4.90 \\
			& OmniAnomaly     & \second{51.17} & 1.61 & 40.09 & 23.48 & 9.68 & 8.47 & \best{82.20} & 14.33 & 24.16 \\
			& TranAD          & 22.63 & 1.65 & 37.28 & 22.36 & 9.57 & 7.75 & 69.97 & 13.51 & 8.41 \\
			& USAD            & 20.99 & 4.07 & \best{61.46} & \second{28.23} & 9.54 & 14.63 & 70.64 & 20.85 & 9.35 \\
			& AnomTrans.      & 2.56 & 0.56 & 8.13 & 7.81 & 15.07 & 4.09 & 22.62 & 6.18 & 1.39 \\
			& TimesNet        & 25.12 & 0.50 & 28.80 & 15.77 & 7.90 & 4.69 & 42.68 & 17.97 & 7.52 \\
			& FITS            & 15.38 & 0.54 & 31.98 & 15.77 & 7.95 & 4.95 & 42.90 & 16.39 & 4.86 \\
			\cmidrule(lr){2-11}
			& TCN-AE          & 39.38 & 3.25 & 36.63 & 19.92 & 10.23 & 13.87 & 68.58 & 24.85 & 58.94 \\
			& TCN-Injection   & 48.37 & \best{23.89} & 46.91 & 27.04 & \second{21.59} & \best{44.79} & 64.45 & 17.26 & 35.68 \\
			& TCN-Supervised  & 41.92 & 1.94 & 42.50 & 20.13 & 11.01 & 9.56 & 72.93 & 22.54 & 59.52 \\
			& TCN-Forecast    & 30.24 & 1.56 & 37.01 & 19.82 & 10.15 & 8.50 & 68.89 & 23.42 & \second{61.61} \\
			\cmidrule(lr){2-11}
			& CAPS            & \best{60.62} & \second{7.58} & \second{57.90} & \best{29.20} & \best{24.27}\mystar & \second{41.53} & \second{80.15} & \best{46.17}\mystar & \best{71.35} \\
			\cmidrule(lr){2-11}
			& CAPS(Diagnosis) & 61.90\mystar & 7.99\mystar & 59.21\mystar & 30.36\mystar & 15.76 & 45.26\mystar & 83.33\mystar & 40.77 & 75.23\mystar \\
			\midrule
			
			\multirow{14}{*}{Standard-F1}
			& LOF             & 30.28 & 1.05 & 24.04 & 12.18 & 4.11 & 4.70 & 56.92 & \second{25.77} & 48.95 \\
			& IForest         & 8.37 & 0.73 & 29.41 & 19.77 & 3.81 & 4.09 & 58.10 & 13.35 & 3.20 \\
			& OmniAnomaly     & \second{47.05} & 1.44 & 28.81 & 23.50 & 0.43 & 5.11 & \second{74.03} & 12.65 & 21.40 \\
			& TranAD          & 34.85 & 1.46 & 27.33 & 22.36 & 2.63 & 4.40 & 60.36 & 11.94 & 5.70 \\
			& USAD            & 30.66 & 3.89 & \best{56.15} & \second{28.24} & 3.41 & 10.74 & 62.91 & 23.87 & 7.21 \\
			& AnomTrans.      & 3.74 & 0.58 & 19.35 & 15.77 & 16.87 & 3.44 & 30.23 & 13.58 & 2.31 \\
			& TimesNet        & 22.52 & 0.45 & 24.98 & 15.77 & 7.90 & 3.84 & 43.90 & 17.58 & 6.97 \\
			& FITS            & 15.81 & 0.49 & 29.00 & 15.77 & 7.95 & 3.86 & 44.27 & 18.08 & 4.58 \\
			\cmidrule(lr){2-11}
			& TCN-AE          & 27.15 & 2.26 & 28.53 & 20.08 & 10.22 & 9.06 & 58.61 & 20.72 & 49.27 \\
			& TCN-Injection   & 46.37 & \best{14.61} & 38.00 & 27.00 & \second{21.42} & \second{35.65} & 60.35 & 16.60 & 31.21 \\
			& TCN-Supervised  & 42.38 & 1.56 & 34.35 & 20.03 & 10.92 & 5.24 & 62.58 & 19.23 & 51.24 \\
			& TCN-Forecast    & 18.65 & 1.30 & 28.56 & 19.77 & 10.16 & 4.94 & 61.32 & 20.56 & \second{54.63} \\
			\cmidrule(lr){2-11}
			& CAPS            & \best{51.63} & \second{4.54} & \second{51.78} & \best{29.37} & \best{24.11}\mystar & \best{37.70} & \best{76.23} & \best{41.24}\mystar & \best{69.58} \\
			\cmidrule(lr){2-11}
			& CAPS(Diagnosis) & 52.14\mystar & 6.84\mystar & 51.82\mystar & 30.30\mystar & 15.78 & 41.24\mystar & 76.78\mystar & 37.22 & 73.55\mystar \\
			\midrule
			
			\multirow{14}{*}{VUS-PR}
			& LOF             & 19.43 & 0.57 & 21.18 & 9.31 & 6.81 & 2.39 & 58.52 & 49.14 & 41.37 \\
			& IForest         & 8.59 & 0.62 & 23.57 & 11.56 & 7.71 & 2.88 & 56.50 & 46.62 & 10.47 \\
			& OmniAnomaly     & 25.35 & 0.64 & 27.17 & 14.32 & 6.20 & 2.40 & \second{74.51} & 45.55 & 29.26 \\
			& TranAD          & 21.61 & 0.64 & 24.82 & 13.04 & 5.75 & 2.25 & 61.63 & 47.33 & 25.78 \\
			& USAD            & 16.58 & 0.75 & \best{55.03} & 18.68 & 4.37 & 8.85 & 58.53 & 56.36 & 14.15 \\
			& AnomTrans.      & 4.60 & 0.43 & 13.11 & 8.84 & 8.29 & 2.12 & 22.67 & 44.74 & 8.00 \\
			& TimesNet        & 17.94 & 0.42 & 20.45 & 7.73 & 4.70 & 2.37 & 44.73 & 57.06 & 25.08 \\
			& FITS            & 13.94 & 0.40 & 25.43 & 7.46 & 5.08 & 2.20 & 38.77 & 56.00 & 18.81 \\
			\cmidrule(lr){2-11}
			& TCN-AE          & 17.27 & 0.87 & 21.56 & 14.95 & 8.99 & 3.63 & 61.04 & 46.30 & 41.03 \\
			& TCN-Injection   & \second{48.69} & \best{3.27} & 34.97 & \best{21.87} & \second{17.50} & \second{26.07} & 60.76 & 49.35 & 42.35 \\
			& TCN-Supervised  & 33.62 & 0.61 & 30.66 & 11.85 & 8.50 & 2.87 & 67.52 & \second{76.61} & \second{44.28} \\
			& TCN-Forecast    & 11.19 & 0.57 & 20.85 & 11.25 & 8.92 & 2.37 & 63.35 & 45.72 & 43.27 \\
			\cmidrule(lr){2-11}
			& CAPS            & \best{49.00} & \second{1.55} & \second{52.35} & \second{19.42} & \best{20.09}\mystar & \best{28.47} & \best{80.52} & \best{86.26}\mystar & \best{83.70} \\
			\cmidrule(lr){2-11}
			& CAPS(Diagnosis) & 50.45\mystar & 1.68\mystar & 52.73\mystar & 21.00\mystar & 13.69 & 31.40\mystar & 81.91\mystar & 81.92 & 84.32\mystar \\
			\bottomrule
		\end{tabular}%
	}
	\vspace*{-1.8mm}
\end{table}

\section{Extended Experiments and Analyses}
\label{appendix:extended_experiments}

\subsection{Supervision Construction and Controlled Comparisons}
\label{appendix:supervision_analysis}

This section provides detailed results and analyses for the
supervision-construction comparisons summarized in
Table~\ref{tab:caps_ablation}.
We examine three complementary aspects: matching generated effects
to their target references, constructing counterparts through
residual parameterization, and comparing with direct source-residual
injection under the same simulated anomaly source.
These comparisons assess how the construction of synthetic
supervision contributes to detection performance.
We additionally examine imbalance-aware label-supervised training
and robustness to contaminated target references.

\subsubsection{Effect of Context--Effect Matching}

A central design of CAPS is that each anomaly effect is conditionally
realized under the same target context on which its counterpart is
constructed. To examine the contribution of this correspondence,
we construct a shuffled variant that preserves the learned generator
and the amount of generated supervision but mismatches each realized
effect with its target reference.

Specifically, for a target reference \(x_i^{\mathrm{tar}}\),
standard CAPS constructs
\(\hat{x}_{i,a}^{(k)}=x_i^{\mathrm{tar}}+\hat r_i^{(k)}\),
where \(\hat r_i^{(k)}\) is generated conditionally on the structure
representation of \(x_i^{\mathrm{tar}}\).
In the shuffled variant, we instead use an effect generated for
another randomly selected reference within the same mode, yielding
\[
\hat{x}_{i,a}^{(k),\mathrm{shuffle}}
=
x_i^{\mathrm{tar}}+\hat r_{\pi(i)}^{(k)},
\qquad
\pi(i)\neq i.
\]
The corresponding anomaly support is shuffled together with the
residual, so each generated counterpart receives the mask associated
with the applied effect.
The number of generated counterparts and the underlying residual
generator are otherwise unchanged.
This is the \emph{Shuffled residuals} variant in
Table~\ref{tab:caps_ablation}.

\begin{table}[H]
	\centering
	\small
	\setlength{\tabcolsep}{4pt}
	\renewcommand{\arraystretch}{1.05}
	\caption{Effect of disrupting context--effect matching in CAPS.
		Higher is better. The better result for each dataset--metric
		pair is bolded.}
	\label{tab:ablation_pairing}
	\resizebox{\linewidth}{!}{%
		\begin{tabular}{ll*{9}{c}}
			\toprule
			Metric & Method
			& IOPS & MGAB & NAB & Power & SED
			& UCR & NEK & TODS & YAHOO \\
			\midrule
			Affiliation-F
			& CAPS
			& \textbf{89.24} & \textbf{69.22} & \textbf{93.25}
			& \textbf{85.98} & \textbf{74.09} & \textbf{88.41}
			& \textbf{86.30} & \textbf{86.85} & \textbf{94.63} \\
			& Shuffled residuals
			& 83.41 & 68.04 & 86.84 & 78.81 & 70.16
			& 81.99 & 81.11 & 75.03 & 88.06 \\
			\midrule
			$\mathrm{F1}_{\mathrm{T}}$
			& CAPS
			& \textbf{60.62} & \textbf{7.58} & \textbf{57.90}
			& \textbf{29.20} & \textbf{24.27} & \textbf{41.53}
			& \textbf{80.15} & \textbf{46.17} & \textbf{71.35} \\
			& Shuffled residuals
			& 53.32 & 1.76 & 52.75 & 20.38 & 22.88
			& 37.18 & 74.98 & 38.15 & 69.50 \\
			\midrule
			Standard-F1
			& CAPS
			& \textbf{51.63} & \textbf{4.54} & \textbf{51.78}
			& \textbf{29.37} & \textbf{24.11} & \textbf{37.70}
			& \textbf{76.23} & \textbf{41.24} & \textbf{69.58} \\
			& Shuffled residuals
			& 46.85 & 1.62 & 46.31 & 20.38 & 22.81
			& 32.29 & 68.40 & 34.12 & 67.26 \\
			\midrule
			VUS-PR
			& CAPS
			& \textbf{49.00} & \textbf{1.55} & \textbf{52.35}
			& \textbf{19.42} & 20.09 & \textbf{28.47}
			& \textbf{80.52} & \textbf{86.26} & \textbf{83.70} \\
			& Shuffled residuals
			& 44.78 & 0.61 & 45.33 & 17.22 & \textbf{20.88}
			& 23.17 & 72.68 & 72.60 & 74.46 \\
			\bottomrule
		\end{tabular}%
	}
\end{table}

Table~\ref{tab:ablation_pairing} provides the per-dataset results
underlying the aggregate comparison in the main text.
Disrupting context--effect matching reduces the average
Affiliation-F, \(\mathrm{F1}_{\mathrm{T}}\), Standard-F1,
and VUS-PR from \(85.33/46.53/42.91/46.82\) to
\(79.27/41.21/37.78/41.30\), respectively.
The degradation occurs on nearly all dataset--metric pairs.
The generator and number of generated counterparts are preserved,
and moving each residual together with its mask maintains their
support correspondence.
These results support the importance of realizing an anomaly effect
for the particular reference on which it is applied: effects generated
under a different temporal context provide less effective supervision
on average, even when the same learned generator is used.

\subsubsection{Effect of Residual-Form Construction}

We next examine the role of constructing counterparts through an
observation-level residual effect.
Standard CAPS generates \(\hat r_i^{(k)}\) conditioned on the
target reference representation and constructs the counterpart as
\(\hat x_{i,a}^{(k)}=x_i^{\mathrm{tar}}+\hat r_i^{(k)}\).
As an ablation, we replace this residual formulation with direct
anomalous-sequence generation while keeping the remaining framework
as unchanged as possible.
Background-preservation supervision outside the intended anomaly
support is retained in the direct-generation variant, so the
comparison primarily changes how the anomalous counterpart is
parameterized.
This corresponds to \emph{w/o residual parameterization} in
Table~\ref{tab:caps_ablation}.

\begin{table}[H]
	\centering
	\small
	\setlength{\tabcolsep}{4pt}
	\renewcommand{\arraystretch}{1.05}
	\caption{Effect of removing residual-form counterpart construction
		in CAPS. Higher is better. The better result for each
		dataset--metric pair is bolded.}
	\label{tab:ablation_residual}
	\resizebox{\linewidth}{!}{%
		\begin{tabular}{ll*{9}{c}}
			\toprule
			Metric & Method
			& IOPS & MGAB & NAB & Power & SED
			& UCR & NEK & TODS & YAHOO \\
			\midrule
			Affiliation-F
			& CAPS
			& \textbf{89.24} & \textbf{69.22} & \textbf{93.25}
			& \textbf{85.98} & 74.09 & \textbf{88.41}
			& \textbf{86.30} & \textbf{86.85} & \textbf{94.63} \\
			& w/o residual parameterization
			& 84.91 & 68.07 & 90.61 & 85.11 & \textbf{77.54}
			& 87.26 & 81.16 & 77.92 & 92.27 \\
			\midrule
			$\mathrm{F1}_{\mathrm{T}}$
			& CAPS
			& \textbf{60.62} & \textbf{7.58} & \textbf{57.90}
			& \textbf{29.20} & \textbf{24.27} & \textbf{41.53}
			& \textbf{80.15} & \textbf{46.17} & \textbf{71.35} \\
			& w/o residual parameterization
			& 44.06 & 1.26 & 54.96 & 24.03 & 19.96
			& 39.59 & 76.53 & 39.91 & 68.37 \\
			\midrule
			Standard-F1
			& CAPS
			& \textbf{51.63} & \textbf{4.54} & \textbf{51.78}
			& \textbf{29.37} & \textbf{24.11} & \textbf{37.70}
			& \textbf{76.23} & \textbf{41.24} & \textbf{69.58} \\
			& w/o residual parameterization
			& 37.98 & 1.15 & 48.91 & 24.02 & 19.87
			& 35.67 & 72.43 & 35.78 & 69.34 \\
			\midrule
			VUS-PR
			& CAPS
			& \textbf{49.00} & \textbf{1.55} & \textbf{52.35}
			& \textbf{19.42} & 20.09 & \textbf{28.47}
			& \textbf{80.52} & \textbf{86.26} & \textbf{83.70} \\
			& w/o residual parameterization
			& 37.39 & 0.57 & 46.35 & 19.31 & \textbf{24.94}
			& 26.80 & 72.90 & 74.85 & 77.32 \\
			\bottomrule
		\end{tabular}%
	}
\end{table}

Table~\ref{tab:ablation_residual} shows that direct
anomalous-sequence generation reduces the four average metrics
from \(85.33/46.53/42.91/46.82\) to
\(82.76/40.96/38.35/42.27\).
The degradation occurs on most dataset--metric pairs, although
direct generation performs better on Affiliation-F and VUS-PR
for SED.
These results support residual-form construction as a useful
reference-preserving parameterization.
By retaining the observed reference explicitly, CAPS allows
the generator to focus on the anomaly-induced observation-level
change while preserving the surrounding temporal background.
This comparison concerns the parameterization of generated
counterparts; it does not assume that the underlying physical
anomaly mechanism is additive.

\subsubsection{Alternative Synthetic Supervision}

We further examine whether the improvement of CAPS can be explained
simply by access to a favorable simulated anomaly source or by exposure
to synthetic anomalous samples. We compare alternative supervision
construction strategies under the same TCN detector backbone,
complementing the aggregate results in
Tables~\ref{tab:overall_results} and~\ref{tab:caps_ablation}.

\paragraph{Direct source-residual injection.}
We first construct a controlled baseline using the same simulated
anomaly source as CAPS. Given a simulated normal--anomalous
correspondence, its observation-level residual is directly applied
to a target reference to construct a synthetic counterpart.
This variant bypasses both anomaly-semantic learning and
target-conditioned realization: it transfers an effect already
realized under a source context, rather than generating an effect
according to the target reference.
We refer to this variant as
\emph{Direct source-residual injection}, consistent with
Table~\ref{tab:caps_ablation}.

\begin{table}[H]
	\centering
	\small
	\setlength{\tabcolsep}{5pt}
	\renewcommand{\arraystretch}{1.05}
	\caption{Comparison with synthetic supervision constructed from
		the same simulated anomaly source. Results are averaged over
		the nine benchmark datasets. Higher is better.}
	\label{tab:same_source_injection}
	\resizebox{\linewidth}{!}{%
		\begin{tabular}{lcccc}
			\toprule
			Method & Affiliation-F & $\mathrm{F1}_{\mathrm{T}}$
			& Standard-F1 & VUS-PR \\
			\midrule
			Direct source-residual injection
			& 81.58 & 43.16 & 39.96 & 42.58 \\
			CAPS
			& \textbf{85.33} & \textbf{46.53}
			& \textbf{42.91} & \textbf{46.82} \\
			\bottomrule
		\end{tabular}%
	}
\end{table}

As shown in Table~\ref{tab:same_source_injection}, CAPS outperforms
direct source-residual injection on all four aggregate metrics.
Because the simulated anomaly source and detector backbone are
shared, the improvement supports the value of learning transferable
anomaly semantics and realizing them under target temporal contexts,
beyond directly reusing source-context effects.

This comparison evaluates the combined contribution of semantic
learning and target-conditioned realization, since both stages are
bypassed together. The context--effect matching and residual-form
comparisons above provide complementary evidence for individual
construction choices. The same direct source-residual injection
baseline is also reported in
Sec.~\ref{appendix:realization_analysis} to contextualize the
comparison of residual-generation mechanisms.

\paragraph{Perturbation-based injection.}
We additionally compare with TCN-Injection in
Table~\ref{tab:detailed_results}, an independent perturbation-based
synthetic-training strategy using the same TCN detector backbone.
TCN-Injection achieves average Affiliation-F,
$\mathrm{F1}_{\mathrm{T}}$, Standard-F1, and VUS-PR scores of
\(80.30/36.66/32.36/33.87\), compared with
\(85.33/46.53/42.91/46.82\) for CAPS.
These results support the importance of how synthetic anomalous
supervision is constructed.

Perturbation-based injection also constructs anomalous samples on
target backgrounds, so this comparison does not isolate
context--effect matching.
Together with direct source-residual injection and the preceding
controlled comparisons, the results support the central design of
CAPS: learning reusable anomaly semantics and adapting their
realization to each target reference provides useful supervision
beyond the synthetic constructions evaluated here.

\subsubsection{Recovered versus Label-Supervised Training}

TCN-Supervised serves as a supervised reference with the same
detector backbone, rather than an optimized upper bound on
supervised TSAD.
It is trained directly on observed target-domain labels using
point-wise BCE and therefore inherits the naturally sparse anomaly
frequency of the training data.
CAPS instead constructs matched reference--counterpart pairs and
assigns equal training weight to the reference and generated
branches.

This construction provides controlled exposure to generated
anomalous counterparts without requiring target anomaly labels.
Equal branch weights, however, do not imply balanced point-wise
classes: the anomaly support \(m\) generally occupies only part of
each generated counterpart, and the remaining timestamps retain
nominal labels.
Any benefit from this training structure should therefore be
distinguished from explicit balancing of anomalous and nominal
timestamps.

To examine whether standard class-imbalance treatments account for
the gap to the supervised reference, we evaluate additional
supervised strategies using the same TCN backbone.
This analysis covers YAHOO, NEK, and TODS, which together contain
283 time-series files and have training anomaly ratios of
\(0.299\%\), \(3.17\%\), and \(3.66\%\), respectively.
Alongside standard BCE, we consider Weighted BCE, Balanced Window
Sampling, and Focal Loss.

\begin{table*}[t]
	\centering
	\small
	\setlength{\tabcolsep}{7pt}
	\renewcommand{\arraystretch}{1.05}
	\caption{Comparison with label-supervised TCN training under
		different class-imbalance treatments. All methods use the
		same TCN detector backbone. Higher is better.}
	\label{tab:supervised_imbalance}
	\begin{tabular}{llcccc}
		\toprule
		Dataset & Method & Affiliation-F & $\mathrm{F1}_{\mathrm{T}}$
		& Standard-F1 & VUS-PR \\
		\midrule
		
		\multirow{5}{*}{YAHOO}
		& TCN-Supervised (BCE)
		& 83.11 & 59.52 & 51.24 & 44.28 \\
		& Weighted BCE
		& 85.10 & 59.90 & 56.90 & 45.70 \\
		& Balanced Window Sampling
		& 83.10 & 37.30 & 34.50 & 39.30 \\
		& Focal Loss
		& 83.70 & 38.50 & 35.70 & 40.30 \\
		& CAPS
		& \textbf{94.63} & \textbf{71.35}
		& \textbf{69.58} & \textbf{83.70} \\
		\midrule
		
		\multirow{5}{*}{NEK}
		& TCN-Supervised (BCE)
		& 85.24 & 72.93 & 62.58 & 67.52 \\
		& Weighted BCE
		& 77.30 & 47.10 & 36.10 & 30.70 \\
		& Balanced Window Sampling
		& 76.50 & 43.50 & 31.80 & 26.80 \\
		& Focal Loss
		& 76.40 & 44.20 & 31.70 & 26.50 \\
		& CAPS
		& \textbf{86.30} & \textbf{80.15}
		& \textbf{76.23} & \textbf{80.52} \\
		\midrule
		
		\multirow{5}{*}{TODS}
		& TCN-Supervised (BCE)
		& 71.75 & 22.54 & 19.23 & 76.61 \\
		& Weighted BCE
		& 74.00 & 23.80 & 21.40 & 76.20 \\
		& Balanced Window Sampling
		& 73.60 & 27.00 & 22.60 & 75.90 \\
		& Focal Loss
		& 73.50 & 24.90 & 21.90 & 75.90 \\
		& CAPS
		& \textbf{86.85} & \textbf{46.17}
		& \textbf{41.24} & \textbf{86.26} \\
		\bottomrule
	\end{tabular}
\end{table*}

Table~\ref{tab:supervised_imbalance} shows that the effects of
imbalance-aware training vary across datasets and metrics.
Weighted BCE improves the YAHOO results over standard BCE,
whereas the standard supervised configuration remains stronger
than the tested alternatives on NEK.
On TODS, all three treatments improve the F1-based metrics, while
slightly reducing VUS-PR.
Thus, these treatments do not uniformly improve supervised
detection performance.

CAPS remains stronger across the reported metrics on all three
datasets under the evaluated protocol.
The tested imbalance-aware strategies therefore do not close the
gap to CAPS, suggesting that the comparison with standard BCE
reflects more than the absence of these particular treatments.
This observation does not establish that recovered supervision
is generally superior to ground-truth labels: a more extensively
optimized supervised pipeline may benefit from different
objectives, sampling policies, or dataset-specific tuning.

The comparison highlights the practical value of the supervision
constructed by CAPS.
Each generated counterpart is contrasted with its own nominal
reference, concentrating the training distinction on an
anomaly-induced change under a shared temporal background.
Pair construction also controls the exposure to generated
anomalous sequences through equal reference and generated branch
weights.
These properties support matched supervision as a useful approach
in label-scarce settings, although the present comparison does
not isolate their individual contributions.

\subsubsection{Robustness to Reference Contamination}

CAPS assumes that target-domain training data are normal-dominated
rather than strictly anomaly-free.
Reference contamination can therefore introduce supervision noise:
an existing anomaly in a reference receives the nominal label
\(\mathbf 0\), and any such anomaly retained outside the generated
support \(m\) is also unmarked in the counterpart.
This reference-label noise is not explicitly modeled by the
coverage and realization terms in
Proposition~\ref{prop:transfer}, motivating an empirical robustness
analysis.

\paragraph{Natural contamination in benchmark data.}
Several benchmarks contain anomalies in their target training
segments.
In YAHOO, 122 of the 259 time-series files contain training-segment
anomalies.
TODS presents a more demanding case, with an average contaminated
training-window ratio of approximately \(74\%\).
This window-level ratio measures how frequently reference windows
contain anomalies; it should be distinguished from the proportion
of anomalous timestamps reported above.
A high contaminated-window ratio can coexist with a relatively low
point-wise anomaly ratio.

CAPS achieves strong results on both datasets in
Table~\ref{tab:detailed_results}, indicating that contaminated
references do not necessarily prevent effective detector learning.
These benchmark observations nevertheless combine contamination
with other dataset characteristics and do not isolate its effect.

\paragraph{Controlled contamination on IOPS.}
We conduct a controlled experiment on IOPS to examine reference
contamination more directly.
Its 17 time-series files provide approximately 21,184 clean
training windows, with a natural training contamination ratio close
to zero, while the test portions provide approximately 26,953
anomalous candidate windows.

For each file, we sample 100 target training windows of length
256 and replace \(0\%\), \(1\%\), \(2\%\), \(5\%\), \(10\%\),
\(20\%\), or \(30\%\) of them with real anomalous windows drawn
from the test portions of other IOPS files.
Anomaly labels are used only to construct this controlled
contamination experiment; the selected windows are treated as
nominal references during CAPS training.
This is a diagnostic stress test, separate from the standard
training protocol without target anomaly labels.
The total number of reference windows remains fixed at 100 per
file across contamination levels, and all remaining representation,
generation, detector-training, and evaluation settings are
unchanged.
Each level is repeated five times, with results averaged over
the 17 files and five runs.
Because this experiment uses a restricted reference set, its
zero-contamination result differs from the main IOPS result.

\begin{table}[H]
	\centering
	\small
	\setlength{\tabcolsep}{6pt}
	\renewcommand{\arraystretch}{1.05}
	\caption{Robustness of CAPS to controlled target-reference
		contamination on IOPS using 100 reference windows per file.
		Results are averaged over 17 time-series files and five runs.
		Higher is better.}
	\label{tab:reference_contamination}
	\begin{tabular}{lcccc}
		\toprule
		Contamination & Affiliation-F & $\mathrm{F1}_{\mathrm{T}}$
		& Standard-F1 & VUS-PR \\
		\midrule
		0\%
		& \textbf{85.74} & \textbf{34.17}
		& \textbf{37.54} & \textbf{34.05} \\
		1\%  & 84.29 & 33.08 & 34.98 & 31.19 \\
		2\%  & 83.94 & 33.05 & 35.48 & 30.77 \\
		5\%  & 84.48 & 31.06 & 32.69 & 29.44 \\
		10\% & 84.53 & 30.83 & 31.65 & 29.46 \\
		20\% & 84.20 & 30.61 & 29.22 & 26.52 \\
		30\% & 82.55 & 28.70 & 33.31 & 26.66 \\
		\bottomrule
	\end{tabular}
\end{table}

Table~\ref{tab:reference_contamination} shows an overall decline
as contamination increases, although individual metrics are not
strictly monotonic.
Affiliation-F remains comparatively stable under mild contamination,
while $\mathrm{F1}_{\mathrm{T}}$, Standard-F1, and VUS-PR show
larger reductions relative to the clean-reference setting.
At \(20\%\)--\(30\%\) contamination, CAPS continues to produce
useful detection results, but the degradation is more apparent,
particularly for VUS-PR.

These results suggest some tolerance to reference-label noise
under the tested contamination protocol, while also demonstrating
its cost.
Clean windows remain the majority at every tested level, so the
experiment does not establish robustness when contaminated
references dominate.
The findings support the use of normal-dominated target data
and identify reference quality as a practical factor affecting
recovered supervision.

\subsection{Disentanglement and Semantic Recombination}
\label{appendix:disentanglement_analysis}

We next examine how paired representation learning supports the
separation of temporal structure and transferable anomaly semantics.
We first evaluate the contribution of within-pair counterfactual
recomposition, and then inspect the learned representations and
their recombination across different reference trajectories.
These analyses complement the aggregate ablation results in
Table~\ref{tab:caps_ablation} and the qualitative examples in
Fig.~\ref{fig:cross_context_main}.

\subsubsection{Effect of Counterfactual Recomposition}

The representation-learning objective combines reconstruction,
background preservation, paired disentanglement, and within-pair
counterfactual recomposition.
In particular, $\mathcal L_{\mathrm{cf}}$ reconstructs an anomalous
sample using its anomaly-semantic code and the structure code of
its matched normal reference:
\[
\mathcal L_{\mathrm{cf}}
=
\mathrm{MSE}
\bigl(
\mathrm{Dec}(z_{n,s},z_a,m),
x_a
\bigr).
\]
Compared with reconstructing $x_a$ from its own structure code
$z_{a,s}$, this operation substitutes the matched normal structure
$z_{n,s}$ while retaining the anomaly-semantic code $z_a$ and
support $m$.
It encourages the two branches to jointly explain the observed
anomalous change: the normal structure provides the reference
context, while the anomaly-semantic code supplies the information
needed to reconstruct its anomalous counterpart.

To assess this contribution, we remove
$\mathcal L_{\mathrm{cf}}$ while retaining
$\mathcal L_{\mathrm{rec}}$,
$\mathcal L_{\mathrm{base}}$, and
$\mathcal L_{\mathrm{dis}}$, with the remaining training procedure
unchanged.
This corresponds to the counterfactual-loss ablation summarized
in Table~\ref{tab:caps_ablation}.

\begin{table}[H]
	\centering
	\small
	\setlength{\tabcolsep}{6pt}
	\renewcommand{\arraystretch}{1.05}
	\caption{Effect of within-pair counterfactual recomposition.
		Results are averaged over the nine benchmark datasets.
		Higher is better.}
	\label{tab:loss_ablation}
	\begin{tabular}{lcccc}
		\toprule
		Variant & Affiliation-F & $\mathrm{F1}_{\mathrm{T}}$
		& Standard-F1 & VUS-PR \\
		\midrule
		w/o $\mathcal L_{\mathrm{cf}}$
		& 83.65 & 42.76 & 39.11 & 43.11 \\
		CAPS
		& \textbf{85.33} & \textbf{46.53}
		& \textbf{42.91} & \textbf{46.82} \\
		\bottomrule
	\end{tabular}
\end{table}

Table~\ref{tab:loss_ablation} shows that removing
$\mathcal L_{\mathrm{cf}}$ reduces performance across all four
aggregate metrics.
The result supports the value of explicitly training the
representations to reconstruct the anomalous outcome after
substituting its matched normal structure.
This requirement goes beyond reconstructing each observation
from its own codes and encourages a representation organization
suited to subsequent semantic transfer.
The training constraint operates within matched simulated pairs;
cross-context recombination is examined separately below.

\subsubsection{Latent Representation Analysis}

\begin{figure}[H]
	\centering
	\includegraphics[width=0.6\linewidth]{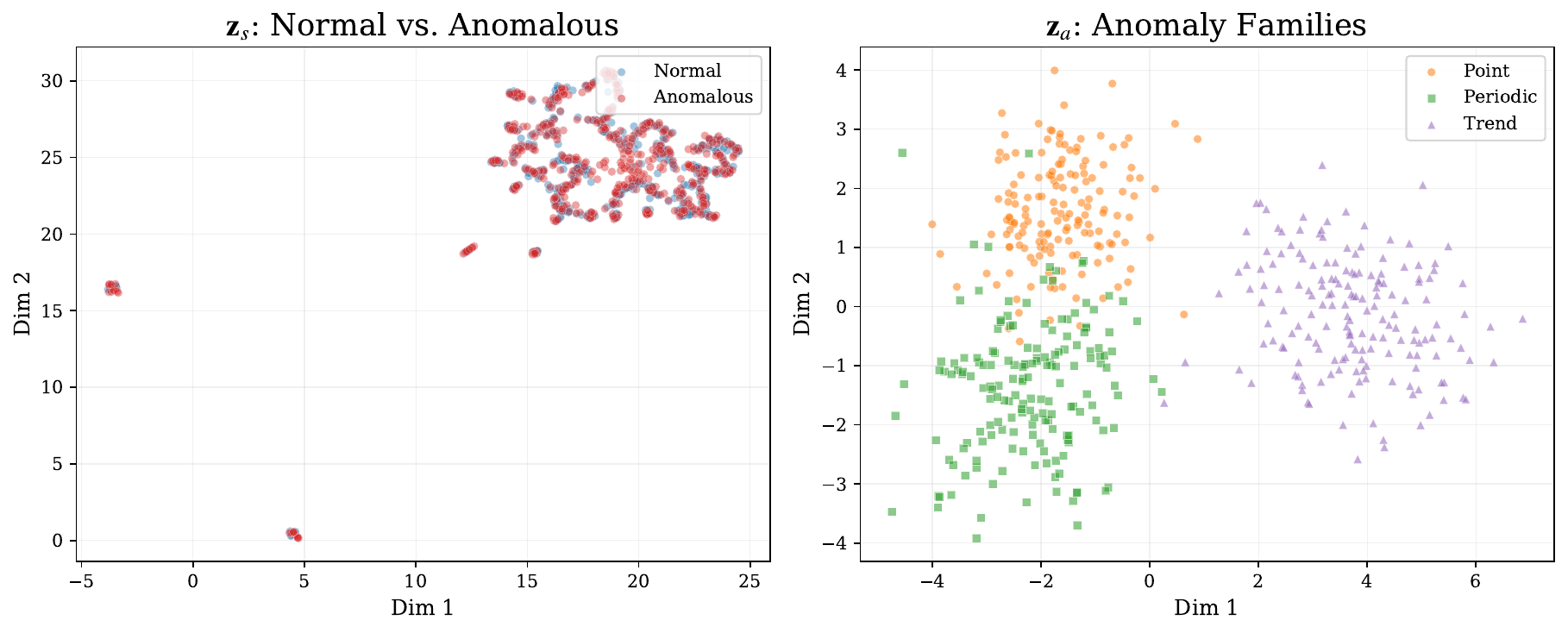}
	\caption{Learned representation spaces.
		\textit{Left}: structure codes $z_s$ colored by
		normal/anomalous status.
		\textit{Right}: anomaly-semantic codes $z_a$ colored
		by simulated anomaly family.}
	\label{fig:disentangle_vis}
\end{figure}

We inspect the learned representations using two-dimensional UMAP
projections of the structure codes $z_s$ and anomaly-semantic
codes $z_a$.
Figure~\ref{fig:disentangle_vis} provides a complementary
visualization to Fig.~\ref{fig:cross_context_main}(a).
The two visualizations use the same trained model but independently
sampled simulated anomaly realizations, resulting in different
projection inputs.
Both exhibit similar qualitative organization of the learned
representations.

Normal and anomalous samples largely overlap in the projected
structure space $z_s$, consistent with reduced sensitivity to
anomaly presence in the structure branch.
The anomaly-semantic space $z_a$, in contrast, exhibits clearer
organization across the point, periodic, and trend families.
This organization is consistent with the use of simulated
families as coarse modes within a shared semantic space,
supporting mode-wise prior estimation and expert specialization.

These projections provide qualitative evidence consistent with
the intended division of representation roles.
They do not by themselves establish statistical independence
between $z_s$ and $z_a$ or semantic invariance across contexts.
The recombination analysis below further examines how the
learned semantic codes interact with different reference
structures during counterpart construction.

\subsubsection{Cross-Context Semantic Recombination}

\begin{figure}[H]
	\centering
	\includegraphics[width=0.6\linewidth]{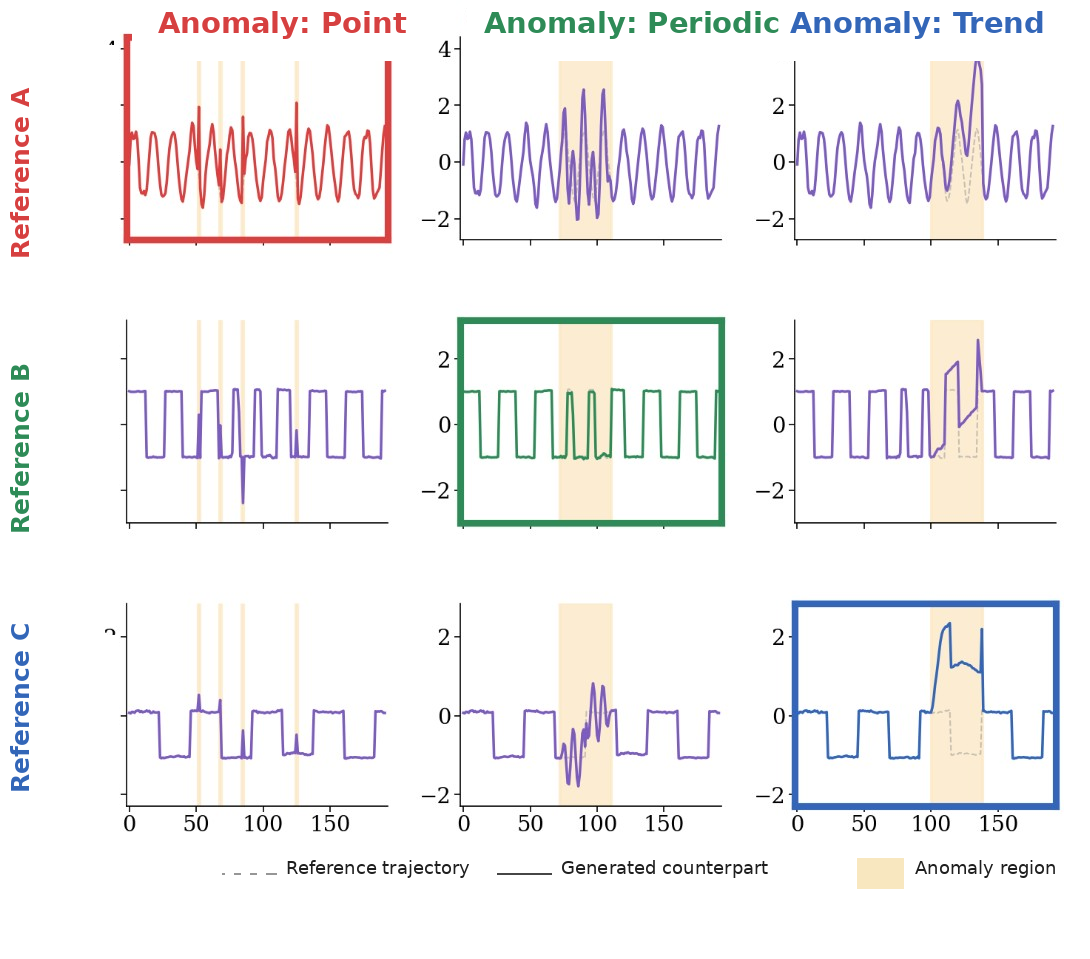}
	\caption{Cross-context semantic recombination.
		Rows use different references, and each column reuses
		a fixed anomaly-semantic code $z_a$.
		Colored diagonal cells mark the original pairings.
		Dashed curves denote references, solid curves denote
		generated counterparts, and shaded regions indicate
		anomaly supports.}
	\label{fig:counterpart_vis}
\end{figure}

We further examine the use of learned anomaly semantics by
recombining fixed semantic codes with different reference
trajectories.
Figure~\ref{fig:counterpart_vis}, corresponding to
Fig.~\ref{fig:cross_context_main}(b), presents a $3\times3$
set of generated counterparts.
Each row uses a different reference trajectory, while each
column reuses one fixed anomaly-semantic code $z_a$.
The Point, Periodic, and Trend column labels identify the coarse
simulated families associated with these codes.
Colored diagonal cells indicate the original reference--semantic
pairings, and off-diagonal cells show cross-context recombinations.

Across the displayed references, each fixed semantic code
produces a recognizable anomaly pattern, while the resulting
counterparts exhibit different observation-level effects.
These examples illustrate the intended distinction between
reusing anomaly semantics and reusing an already realized
source residual.
In CAPS, the semantic code supplies transferable anomaly
information, while the generator also receives the reference
structure and intended support to construct the counterpart.

The generated trajectories also largely preserve the reference
background outside the intended anomaly support.
Together with the counterfactual-loss ablation and latent
visualization, these observations support the central role of
paired representation learning in supervision recovery:
learning anomaly information that can be recombined with
reference structure and subsequently realized as
context-anchored training counterparts.
The cross-context examples are qualitative diagnostics of the
learned model, rather than additional training constraints.

\subsection{Semantic Coverage and Mode Organization}
\label{appendix:coverage_analysis}

We next examine the coverage and organization of the
simulation-induced semantic prior.
CAPS uses coarse source-side modes to organize a shared continuous
anomaly-semantic space and support expert specialization.
These modes do not constitute an exhaustive taxonomy of target
anomalies, which may exhibit mixed, context-dependent, or previously
unseen patterns.
The transferable representation is the continuous semantic code,
while the mode index provides a coarse organization for prior
estimation and generation.
We study three aspects of this design: sensitivity to reduced
coverage, alternative mode organizations, and targeted mode
selection with limited target anomaly guidance.

\subsubsection{Coverage Stress Test}

The usefulness of simulation-derived supervision depends partly
on whether the semantic prior captures anomaly variation relevant
to the target domain.
We examine this dependence by removing the source-side mode most
strongly associated with each target dataset.

Given a small known-anomaly set used only for this diagnostic
analysis, the router produces normalized mode-association scores
$q_k(x)$.
We compute
\[
\bar q_k
=
\frac{1}{|\mathcal D_{\mathrm{known}}|}
\sum_{x\in\mathcal D_{\mathrm{known}}}q_k(x),
\qquad
k^\star=\arg\max_k\bar q_k.
\]
During target counterpart construction, we exclude the prior
component and residual expert associated with $k^\star$.
Additional counterparts are sampled from the remaining modes so
that the total number of generated counterparts per reference is
unchanged.
The known anomalies identify the mode to exclude; they are not
supplied as labeled examples for detector training.

\begin{table}[H]
	\centering
	\small
	\setlength{\tabcolsep}{5pt}
	\renewcommand{\arraystretch}{1.05}
	\caption{Semantic-coverage stress test.
		The source-side mode with the highest target association
		is excluded during counterpart generation while the total
		number of counterparts is preserved.
		Results are averaged over the nine benchmark datasets.
		Higher is better.}
	\label{tab:coverage_stress}
	\resizebox{\linewidth}{!}{%
		\begin{tabular}{lcccc}
			\toprule
			Method & Affiliation-F & $\mathrm{F1}_{\mathrm{T}}$
			& Standard-F1 & VUS-PR \\
			\midrule
			CAPS w/o highest-associated mode
			& 79.31 & 38.42 & 37.81 & 39.87 \\
			CAPS
			& \textbf{85.33} & \textbf{46.53}
			& \textbf{42.91} & \textbf{46.82} \\
			\bottomrule
		\end{tabular}%
	}
\end{table}

Table~\ref{tab:coverage_stress} shows lower aggregate performance
after removing the highest-associated mode.
Because the number of generated counterparts is preserved, the
degradation supports the importance of the anomaly variation
available for supervision construction beyond the amount of
synthetic training data.
The remaining modes still provide useful supervision, although
this does not establish coverage of arbitrary unseen anomaly
patterns.

The excluded mode is defined by router association, rather than
a ground-truth anomaly category for the target dataset.
Real datasets may contain heterogeneous patterns that are only
partially represented by any single source-side family.
The stress test therefore examines the consequence of restricting
an associated region of the source prior.
Its results are consistent with the role of coverage discussed in
Sec.~\ref{sec:analysis}, without treating router associations as
a direct measurement of the theoretical coverage discrepancy.

\subsubsection{Mode Construction and Granularity}
\label{appendix:sensitivity_k}

We next examine the dependence of CAPS on its default
knowledge-based mode organization.
This organization uses $K=3$ coarse modes corresponding to point,
periodic, and trend variations in the simulated anomaly operations.
All anomaly-semantic codes remain in the shared continuous space
$\mathcal Z_A$.
The modes organize prior estimation and expert specialization;
they do not define separate latent spaces or assume an additive
physical decomposition of target anomalies.

As a data-driven alternative, we freeze the learned representation
modules and organize the simulated anomaly-semantic codes $z_a$
directly in the latent space.
We evaluate $K\in\{1,3,5,7\}$, where $K=1$ uses a single mode
and $K>1$ uses $K$-means clustering.
The resulting assignments replace the predefined source-family
labels in prior estimation, expert training, and mask-pool
construction.
When Diagnosis is used, they also provide the corresponding
router supervision, as described in
Appendix~\ref{appendix:mode_details}.

\begin{table}[H]
	\centering
	\small
	\setlength{\tabcolsep}{5.5pt}
	\renewcommand{\arraystretch}{1.05}
	\caption{Comparison of semantic-mode organization and
		granularity.
		Results are averaged over the nine benchmark datasets.
		Higher is better.}
	\label{tab:mode_construction}
	\begin{tabular}{llcccc}
		\toprule
		Organization & $K$ & Affiliation-F
		& $\mathrm{F1}_{\mathrm{T}}$ & Standard-F1 & VUS-PR \\
		\midrule
		Data-driven & 1
		& 82.34 & 39.92 & 37.23 & 41.36 \\
		Data-driven & 3
		& 84.98 & 43.75 & 40.12 & 45.76 \\
		Data-driven & 5
		& 83.10 & 44.04 & 40.43 & 44.76 \\
		Data-driven & 7
		& 81.61 & 42.38 & 38.78 & 43.38 \\
		\midrule
		Knowledge-based & 3
		& \textbf{85.33} & \textbf{46.53}
		& \textbf{42.91} & \textbf{46.82} \\
		\bottomrule
	\end{tabular}
\end{table}

Table~\ref{tab:mode_construction} shows that data-driven
organizations also yield effective detectors.
In particular, clustering with $K=3$ provides a competitive
alternative, indicating that the learned semantic space supports
mode organization without retaining the predefined family
assignments in the subsequent generation stage.

At the same $K=3$, the knowledge-based organization achieves
higher aggregate performance across all four metrics.
This comparison holds the number of experts and generated
counterparts per reference constant, providing a more direct
assessment of the organization strategy.
The result supports the usefulness of coarse temporal
characteristics as an inductive bias for specialization.
One possible explanation is that these assignments group
compatible simulated variations for residual generation, although
the present comparison does not directly measure interference
between experts.

Among the data-driven variants, increasing $K$ does not produce
uniform improvements.
Relative to $K=3$, $K=5$ slightly improves
$\mathrm{F1}_{\mathrm{T}}$ and Standard-F1 but reduces
Affiliation-F and VUS-PR, while $K=7$ performs worse on all four.
Changing $K$ also changes the number of residual experts and,
under the standard generation protocol, the number of generated
counterparts per reference.
These results therefore compare complete mode-organization
configurations rather than isolating specialization granularity.

The findings support a coarse organization of continuous anomaly
semantics without requiring target anomalies to be assigned to
predefined categories.
They do not imply that continuous latent variation alone
guarantees coverage of anomaly patterns absent from simulation.
The coverage stress test above highlights this remaining
dependence on the source prior.

\subsubsection{Diagnosis and Targeted Mode Selection}

The mode organization also provides an optional interface for
incorporating limited target anomaly guidance.
Standard CAPS uses all source-side modes without target anomaly
labels, preserving broad semantic coverage when the relevant
target variation is unknown.

CAPS(Diagnosis) uses known target anomalies to select associated
source-side modes for counterpart generation.
In this diagnostic experiment, we randomly sample $10\%$ of
anomalous windows from the target test partition to form
$\mathcal D_{\mathrm{known}}$.
The router produces normalized association scores $q_k(x)$,
which are averaged as
\[
\bar q_k
=
\frac{1}{|\mathcal D_{\mathrm{known}}|}
\sum_{x\in\mathcal D_{\mathrm{known}}}q_k(x).
\]
We sort the modes such that
\[
\bar q_{k_{(1)}}\ge\cdots\ge\bar q_{k_{(K)}}
\]
and retain the smallest subset whose cumulative association
reaches $\rho$:
\[
J_\rho
=
\min\left\{
J:\sum_{j=1}^{J}\bar q_{k_{(j)}}\ge\rho
\right\},
\qquad
\mathcal K_\rho
=
\{k_{(1)},\ldots,k_{(J_\rho)}\}.
\]
We use $\rho=0.8$.
Counterparts are generated uniformly from the retained modes;
the association scores determine mode inclusion rather than
generation mixture weights.

The known anomalies guide supervision construction through mode
selection but are not directly included as labeled examples in
detector training.
Evaluation uses the full test partition, including the sampled
known-anomaly windows.
Accordingly, these results describe a diagnostic extension under
partial test-set guidance, rather than the standard setting
without target anomaly labels or an independent evaluation
entirely on unseen test windows.

\begin{figure}[H]
	\centering
	\includegraphics[width=0.5\linewidth]{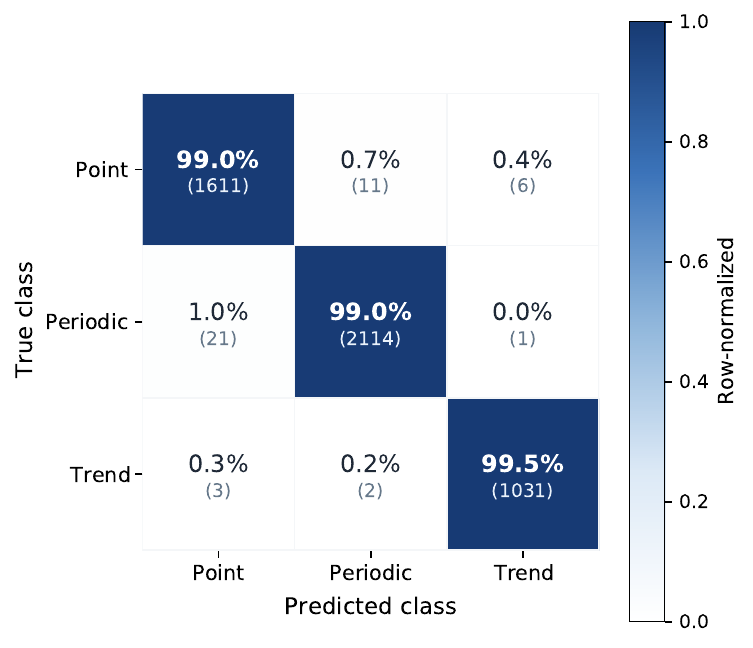}
	\caption{Router associations on simulated samples.
		Rows denote source-side modes and columns denote
		router-associated modes.
		Values are row-normalized percentages, with sample counts
		in parentheses.}
	\label{fig:routing_vis}
\end{figure}

Figure~\ref{fig:routing_vis} shows that the reported router
associations largely agree with the simulated mode assignments,
with most entries concentrated on the diagonal.
This supports consistency with the source-side organization on
the displayed samples.
It does not establish that real target anomalies belong to the
same discrete categories.
For target data, the scores serve as relative associations with
the available source modes.

The detailed results in Table~\ref{tab:detailed_results} show that
CAPS(Diagnosis) improves over standard CAPS on many
dataset--metric pairs, but not universally.
Representative known anomalies may help focus generation on
relevant semantic variation.
Conversely, a small known subset may omit anomaly patterns
present elsewhere in a heterogeneous dataset, causing mode
selection to discard useful coverage.
Under the one-counterpart-per-selected-mode protocol, selection
also changes the amount of generated supervision, so the
comparison evaluates the targeted-generation procedure as a
whole.

Diagnosis therefore provides a way to adapt generation when
limited target guidance is available, without guaranteeing a
performance improvement.
Standard CAPS retains all modes to preserve broader coverage
when such guidance is unavailable.

\subsection{Context-Conditioned Realization}
\label{appendix:realization_analysis}

Having examined semantic representation and coverage, we next
study how anomaly semantics are realized on target references.
In CAPS, $z_a$ provides anomaly-semantic information, while the
reference structure $z_s$ and intended support $m$ also condition
the generated observation-level effect.
This stage connects the simulation-induced semantic prior to
context-anchored target supervision.
We examine alternative residual-generation mechanisms and
compare generated counterparts with real target anomalies.

\subsubsection{Alternative Realization Mechanisms}

We examine how the choice of realization mechanism affects the
supervision constructed for detector learning.
Standard CAPS uses conditional diffusion to generate residual
effects from anomaly semantics, reference structure, and anomaly
support.
We compare this implementation with a VAE-style conditional
residual generator.
We additionally report \emph{Direct source-residual injection},
the same baseline used in Table~\ref{tab:same_source_injection},
which applies simulated source residuals directly to target
references.

The VAE comparison replaces the conditional generation mechanism,
whereas direct source-residual injection bypasses both semantic
learning and target-conditioned realization.
The latter therefore provides a reference for the complete
transfer procedure rather than an isolated generator ablation.
All variants use the same TCN detector backbone and evaluation
protocol.

\begin{table}[H]
	\centering
	\small
	\setlength{\tabcolsep}{5pt}
	\renewcommand{\arraystretch}{1.05}
	\caption{Comparison of anomaly-effect construction strategies.
		Results are averaged over the nine benchmark datasets.
		Higher is better.}
	\label{tab:realization_mechanisms}
	\resizebox{\linewidth}{!}{%
		\begin{tabular}{lcccc}
			\toprule
			Strategy
			& Affiliation-F
			& $\mathrm{F1}_{\mathrm{T}}$
			& Standard-F1
			& VUS-PR \\
			\midrule
			Direct source-residual injection
			& 81.58 & 43.16 & 39.96 & 42.58 \\
			VAE residual generator
			& 83.28 & 42.32 & 38.82 & 42.63 \\
			CAPS (diffusion)
			& \textbf{85.33} & \textbf{46.53}
			& \textbf{42.91} & \textbf{46.82} \\
			\bottomrule
		\end{tabular}%
	}
\end{table}

Table~\ref{tab:realization_mechanisms} shows that conditional
diffusion achieves the highest aggregate performance across all
four metrics.
The two alternatives exhibit mixed relative performance:
the VAE generator performs better on Affiliation-F and slightly
better on VUS-PR, while direct source-residual injection performs
better on $\mathrm{F1}_{\mathrm{T}}$ and Standard-F1.
Thus, introducing a learned conditional generator alone does not
uniformly improve over direct injection; its implementation
also matters for the resulting detector supervision.

The comparison with the VAE supports the diffusion implementation
used in CAPS, while the comparison with direct source-residual
injection supports the combined semantic-learning and
target-conditioned realization procedure.
These results do not establish diffusion as a necessary component
of supervision recovery.
Among the evaluated implementations, however, it provides the
most effective residual generator for downstream detection.
Its computational cost is examined in the efficiency analysis
below; generation is used for offline supervision construction,
while inference requires only the trained detector.

\subsubsection{Generated-Counterpart Diagnostics against Real Anomalies}

Downstream detection results do not directly characterize the
relationship between generated counterparts and observed target
anomalies.
We therefore complement the detector comparisons with two
diagnostics in detector-score and feature spaces.

We first train a detector using standard CAPS supervision.
For held-out normal references not used in detector training,
we construct equal numbers of counterparts using rule-based
injection, direct source-residual injection, and CAPS.
The three strategies share the same reference windows, anomaly
supports, and sample counts.
We then pass the generated counterparts and held-out real target
anomalies through the same fixed detector.

\emph{Score} measures the Wasserstein distance between the
detector-score distributions of generated and real anomalies.
\emph{Feature} measures the average distance from each generated
counterpart to its five nearest real-anomaly neighbors in the
penultimate-layer feature space.
Lower values indicate greater proximity under the corresponding
detector-derived representation.
Real anomaly labels are used for this diagnostic evaluation,
not for standard CAPS detector training.

\begin{table}[H]
	\centering
	\small
	\setlength{\tabcolsep}{8pt}
	\renewcommand{\arraystretch}{1.05}
	\caption{Detector-oriented diagnostics against real target
		anomalies, averaged over the nine benchmark datasets.
		Score measures Wasserstein distance between anomaly-score
		distributions; Feature measures average generated-to-real
		$5$-NN distance in the penultimate-layer feature space.
		Lower is better.}
	\label{tab:counterpart_diagnostics}
	\begin{tabular}{lcc}
		\toprule
		Positive source & Score $\downarrow$ & Feature $\downarrow$ \\
		\midrule
		Rule-based injection
		& 0.4531 & 0.6218 \\
		Direct source-residual injection
		& 0.4783 & 0.6712 \\
		CAPS
		& \textbf{0.3990} & \textbf{0.6123} \\
		\bottomrule
	\end{tabular}
\end{table}

Table~\ref{tab:counterpart_diagnostics} shows that CAPS has the
lowest average discrepancy under both diagnostics.
Its generated counterparts are closer to real anomalies in
detector-score distribution and in the selected feature-space
measure.
The feature-distance improvement over rule-based injection is
modest, so these averages should not be interpreted as establishing
a large or uniform advantage across datasets.

These observations complement the supervision-construction
comparisons in Appendix~\ref{appendix:supervision_analysis}.
Direct source-residual injection reuses effects already realized
under simulated contexts, while rule-based injection constructs
effects through predefined perturbations.
CAPS instead learns anomaly-semantic representations and
conditions their realization on the target reference.
The smaller reported distances are consistent with this procedure
producing supervision relevant to the real anomalies encountered
by the detector.

The interpretation remains specific to the diagnostic setting.
Because the detector is trained on CAPS-generated supervision,
its scores and features are not independent of the method being
evaluated and may favor aspects of CAPS counterparts.
Moreover, generated-to-real nearest-neighbor distance measures
local proximity rather than coverage of the full real-anomaly
distribution.
These diagnostics therefore provide complementary observations
in a downstream representation space; they do not establish
recovery of true counterfactual outcomes or complete agreement
with the distribution of real target anomalies.

\subsection{Robustness, Scalability, and Efficiency}
\label{appendix:robustness}

We finally examine the sensitivity of CAPS to detector architecture
and semantic-prior choices, its behavior under different amounts
of simulated training data, and its computational cost.
These analyses assess the flexibility of recovered supervision
and distinguish the cost of offline representation learning and
counterpart generation from that of detector inference.

\subsubsection{Detector Architecture Variants}

The main experiments use a lightweight TCN detector.
To examine the applicability of recovered supervision to a
different architecture, we replace the TCN with a Transformer
while retaining the supervision-construction procedure and
training protocol.

\begin{table}[H]
	\centering
	\small
	\setlength{\tabcolsep}{6pt}
	\renewcommand{\arraystretch}{1.05}
	\caption{Comparison of CAPS under different detector architectures.
		Results are averaged over the nine benchmark datasets.
		Higher is better.}
	\label{tab:backbone_results}
	\begin{tabular}{lcc}
		\toprule
		Metric & CAPS (TCN) & CAPS (Transformer) \\
		\midrule
		Affiliation-F
		& \textbf{85.33} & 81.10 \\
		$\mathrm{F1}_{\mathrm{T}}$
		& \textbf{46.53} & 42.12 \\
		Standard-F1
		& \textbf{42.91} & 40.37 \\
		VUS-PR
		& \textbf{46.82} & 43.28 \\
		\bottomrule
	\end{tabular}
\end{table}

Table~\ref{tab:backbone_results} shows that the recovered pairs
can also support training a Transformer detector, extending their
use beyond the default TCN.
The TCN achieves higher aggregate performance across all four
metrics under the evaluated protocol and is retained as the
default.
The comparison supports flexibility in detector choice, while
also showing that the downstream architecture affects how
effectively the recovered supervision is used.

\subsubsection{Prior Sensitivity}
\label{appendix:prior}

CAPS uses a parametric prior to sample learned anomaly semantics
during target-domain counterpart generation.
The default implementation fits a diagonal Gaussian within each
coarse mode.
We compare this choice with an isotropic Gaussian, a
full-covariance Gaussian, and a Student-\emph{t} distribution,
while keeping the mode organization and remaining settings
unchanged.

\begin{table}[H]
	\centering
	\small
	\setlength{\tabcolsep}{5pt}
	\renewcommand{\arraystretch}{1.05}
	\caption{Sensitivity to the parametric form of the mode-wise
		anomaly-semantic prior.
		Results are averaged over the nine benchmark datasets.
		Higher is better.}
	\label{tab:prior_results}
	\begin{tabular}{lcccc}
		\toprule
		Prior
		& Affiliation-F
		& $\mathrm{F1}_{\mathrm{T}}$
		& Standard-F1
		& VUS-PR \\
		\midrule
		Isotropic Gaussian
		& 84.97 & 45.88 & 42.36 & 46.21 \\
		Diagonal Gaussian
		& 85.33 & 46.53 & \textbf{42.91} & 46.82 \\
		Full-covariance Gaussian
		& \textbf{85.40} & \textbf{47.11} & 42.74 & 46.70 \\
		Student-\emph{t}
		& 85.21 & 46.47 & 42.68 & \textbf{46.95} \\
		\bottomrule
	\end{tabular}
\end{table}

Table~\ref{tab:prior_results} shows similar aggregate performance
across the tested prior families, with no single choice achieving
the best result on every metric.
The full-covariance Gaussian performs best on Affiliation-F and
$\mathrm{F1}_{\mathrm{T}}$, the diagonal Gaussian on Standard-F1,
and the Student-\emph{t} prior on VUS-PR.
The diagonal Gaussian therefore offers a competitive default
using only per-dimension means and variances.

These results suggest limited sensitivity to the tested
within-mode parameterizations under the current semantic
organization.
They concern how the learned codes are approximated for
sampling; they do not establish that changing the prior family
can compensate for anomaly variation absent from the simulated
source.

\subsubsection{Different Amounts of Simulated Data}
\label{appendix:data_scale}

The representation and generation stages are trained using
simulated normal--anomalous correspondences.
To examine their dependence on the amount of source supervision,
we compare the default 48{,}000 simulated pairs with smaller
sets of 5{,}000, 10{,}000, and 20{,}000 pairs, keeping the
remaining settings unchanged.

\begin{table}[H]
	\centering
	\small
	\setlength{\tabcolsep}{6pt}
	\renewcommand{\arraystretch}{1.05}
	\caption{Performance under different amounts of simulated
		training data, averaged over the nine benchmark datasets.
		Higher is better.}
	\label{tab:data_scale}
	\begin{tabular}{lcccc}
		\toprule
		Metric & 5,000 & 10,000 & 20,000 & 48,000 \\
		\midrule
		Affiliation-F
		& 77.66 & 78.93 & 82.15 & \textbf{85.33} \\
		$\mathrm{F1}_{\mathrm{T}}$
		& 44.36 & \textbf{47.24} & 46.21 & 46.53 \\
		Standard-F1
		& 41.05 & \textbf{43.52} & 42.78 & 42.91 \\
		VUS-PR
		& 39.68 & 42.40 & 44.79 & \textbf{46.82} \\
		\bottomrule
	\end{tabular}
\end{table}

Table~\ref{tab:data_scale} shows different responses across
metrics.
Affiliation-F and VUS-PR improve consistently as the number of
simulated pairs increases.
In contrast, $\mathrm{F1}_{\mathrm{T}}$ and Standard-F1 reach
their highest values at 10{,}000 pairs and remain comparatively
close at the larger scales.
Thus, additional simulation benefits some aspects of detection
more clearly than others.

Smaller simulated sets already support useful detector training.
We retain 48{,}000 pairs as the default because this configuration
achieves the strongest Affiliation-F and VUS-PR while maintaining
comparable F1 performance.
The results also identify smaller source sets as practical
alternatives when offline training resources are limited.

Because the epoch counts and batch size remain fixed, larger
source sets also entail more optimization updates.
The comparison therefore measures the effect of increasing
simulation scale under the adopted training schedule, rather
than isolating data quantity at a fixed optimization budget.
The associated training costs are reported below.

\subsubsection{Computational Efficiency}
\label{appendix:efficiency}

The representation and generation modules are trained offline.
Afterward, they are reused to construct counterparts for target
references, and a separate detector is trained on the recovered
supervision.
At deployment, anomaly scoring requires only this detector;
the representation encoder, semantic prior, and diffusion
generator are not part of the inference path.

Table~\ref{tab:data_scale_efficiency} reports simulated-domain
training time at different source-data scales, together with
per-pair generation and test-time detection costs.
These measurements distinguish the cost of constructing
supervision from the cost of scoring unseen sequences.
Target-domain detector fitting is a separate stage and is not
itemized in this table.

\begin{table}[H]
	\centering
	\small
	\setlength{\tabcolsep}{6pt}
	\renewcommand{\arraystretch}{1.05}
	\caption{Computational cost under different amounts of
		simulated training data.
		Pair generation is performed offline; test-time detection
		uses only the trained TCN.}
	\label{tab:data_scale_efficiency}
	\resizebox{\linewidth}{!}{%
		\begin{tabular}{lccc}
			\toprule
			Number of simulated pairs
			& Simulated-domain training
			& Pair generation
			& Test-time detection \\
			\midrule
			5,000
			& 23.6 min & 50 ms / pair & 2 ms / sequence \\
			10,000
			& 48.5 min & 50 ms / pair & 2 ms / sequence \\
			20,000
			& 95.1 min & 50 ms / pair & 2 ms / sequence \\
			48,000
			& 180.5 min & 50 ms / pair & 2 ms / sequence \\
			\bottomrule
		\end{tabular}%
	}
\end{table}

Simulated-domain training time increases with the amount of
source supervision.
Under the fixed model and sampling configuration, the reported
generation cost remains approximately 50\,ms per pair, while
the deployed detector takes approximately 2\,ms per test
sequence.
The total cost of counterpart construction additionally depends
on the number of target references and generated counterparts.

These measurements illustrate the computational separation
between supervision recovery and deployment.
CAPS incurs representation-learning and generation costs before
detector deployment, allowing the conditional generator to support
training without adding diffusion sampling to test-time scoring.

\section{Multivariate Extension}
\label{appendix:multivariate}

The main formulation of CAPS focuses on temporal anomaly effects
without explicitly modeling dependencies among variables.
We examine its extension to multivariate series through a
variable-wise supervision-recovery procedure.
This supplementary experiment evaluates the usefulness of the
temporal supervision across variables, while leaving explicit
dependency modeling for future work.

We evaluate on MSL, PSM, SMAP, and SMD.
Table~\ref{tab:multi_datasets} summarizes the multivariate
sequences included in these experiments.

\begin{table}[H]
	\centering
	\small
	\setlength{\tabcolsep}{5pt}
	\renewcommand{\arraystretch}{1.05}
	\caption{Multivariate datasets used in the supplementary
		experiments.
		\#TS denotes the number of evaluated multivariate sequences,
		\#Dim the number of input variables per sequence, and AR
		the anomaly ratio.}
	\label{tab:multi_datasets}
	\begin{tabular}{llcccc}
		\toprule
		Name & Domain & \#TS & \#Dim & Avg. Length & AR (\%) \\
		\midrule
		MSL
		& Space & 16 & 55 & 3119.4 & 5.1 \\
		PSM
		& Sensor & 1 & 25 & 217624.0 & 11.2 \\
		SMAP
		& Space & 27 & 25 & 7855.9 & 2.9 \\
		SMD
		& Server & 22 & 38 & 25466.4 & 3.8 \\
		\bottomrule
	\end{tabular}
\end{table}

For CAPS, we apply the univariate pipeline independently to each
variable.
The pretrained representation and generation modules are shared
across variables, while a separate TCN detector is trained for
each variable without detector parameter sharing.
The resulting variable-wise anomaly scores are aggregated at
each timestamp to obtain a multivariate score.

We evaluate two aggregation strategies.
\textbf{CAPS (Max)} takes the maximum variable-wise score,
whereas \textbf{CAPS (Top5-mean)} averages the five largest scores.
Neither strategy introduces a module for learning cross-variable
interactions.
Because the extension uses one detector per variable, its
parameter count and computational cost differ from those of
the single-detector univariate setting.

We compare with the multivariate implementations of LOF,
IForest, OmniAnomaly, TranAD, USAD, Anomaly Transformer,
TimesNet, and FITS under the same data splits and evaluation
protocol.

\begin{table}[H]
	\centering
	\small
	\setlength{\tabcolsep}{6pt}
	\renewcommand{\arraystretch}{1.05}
	\caption{Results on multivariate anomaly detection benchmarks.
		Higher is better.
		Best and second-best results within each dataset--metric
		pair are shown in \textbf{bold} and
		\underline{underline}, respectively.}
	\label{tab:multi_results}
	\begin{tabular}{llcccc}
		\toprule
		Metric & Model & MSL & PSM & SMAP & SMD \\
		\midrule
		
		\multirow{10}{*}{Affiliation-F}
		& LOF
		& 84.35 & 61.98 & 63.32 & 64.13 \\
		& IForest
		& 63.36 & 63.78 & 59.96 & 69.71 \\
		& OmniAnomaly
		& 83.15 & 58.17 & 91.38 & 85.82 \\
		& TranAD
		& 79.91 & 73.83 & 87.39 & \textbf{92.20} \\
		& USAD
		& 81.86 & 57.86 & 87.25 & 85.09 \\
		& AnomTrans.
		& 74.38 & 66.04 & 74.82 & 73.44 \\
		& TimesNet
		& 84.38 & 69.43 & 86.69 & 86.90 \\
		& FITS
		& 82.32 & 75.55 & 81.41 & \underline{89.53} \\
		\cmidrule(lr){2-6}
		& CAPS (Max)
		& \underline{86.64} & \textbf{82.71}
		& \textbf{92.87} & 89.37 \\
		& CAPS (Top5-mean)
		& \textbf{88.51} & \underline{79.68}
		& \underline{91.78} & 88.62 \\
		
		\midrule
		
		\multirow{10}{*}{$\mathrm{F1}_{\mathrm{T}}$}
		& LOF
		& 38.97 & 25.58 & 21.81 & 10.13 \\
		& IForest
		& 20.73 & 25.39 & 14.32 & 16.20 \\
		& OmniAnomaly
		& 49.36 & 30.42 & \underline{46.63} & 51.84 \\
		& TranAD
		& 39.42 & 25.49 & 29.12 & 37.98 \\
		& USAD
		& 48.71 & 28.96 & 43.94 & 50.41 \\
		& AnomTrans.
		& 5.91 & 19.55 & 7.18 & 4.82 \\
		& TimesNet
		& 26.95 & 20.55 & 23.47 & 34.91 \\
		& FITS
		& 23.86 & 20.15 & 21.47 & 29.92 \\
		\cmidrule(lr){2-6}
		& CAPS (Max)
		& \textbf{55.07} & \textbf{34.09}
		& \textbf{47.61} & \underline{54.23} \\
		& CAPS (Top5-mean)
		& \underline{54.27} & \underline{32.22}
		& 44.37 & \textbf{57.11} \\
		
		\midrule
		
		\multirow{10}{*}{Standard-F1}
		& LOF
		& 30.65 & 18.80 & 18.70 & 8.41 \\
		& IForest
		& 14.68 & 24.15 & 13.61 & 16.89 \\
		& OmniAnomaly
		& 39.10 & \underline{30.43}
		& 40.50 & \underline{57.06} \\
		& TranAD
		& 29.60 & 25.63 & 25.11 & 43.99 \\
		& USAD
		& 38.71 & 28.41 & 38.66 & 53.06 \\
		& AnomTrans.
		& 10.65 & 23.37 & 8.02 & 9.86 \\
		& TimesNet
		& 20.13 & 20.15 & 23.37 & 39.48 \\
		& FITS
		& 16.96 & 20.15 & 19.99 & 34.18 \\
		\cmidrule(lr){2-6}
		& CAPS (Max)
		& \textbf{46.04} & \textbf{31.18}
		& \textbf{43.33} & 52.39 \\
		& CAPS (Top5-mean)
		& \underline{40.04} & 25.40
		& \underline{40.84} & \textbf{57.83} \\
		
		\midrule
		
		\multirow{10}{*}{VUS-PR}
		& LOF
		& 24.67 & 13.58 & 10.59 & 4.40 \\
		& IForest
		& 11.29 & 15.85 & 7.55 & 8.88 \\
		& OmniAnomaly
		& 31.57 & 18.58 & 28.07 & 37.44 \\
		& TranAD
		& 14.78 & 16.49 & 13.37 & 28.34 \\
		& USAD
		& 29.95 & 17.59 & 26.37 & 34.53 \\
		& AnomTrans.
		& 8.15 & 16.90 & 4.39 & 5.68 \\
		& TimesNet
		& 14.23 & 15.22 & 15.38 & 34.58 \\
		& FITS
		& 13.98 & 14.02 & 14.85 & 30.32 \\
		\cmidrule(lr){2-6}
		& CAPS (Max)
		& \textbf{40.15} & \textbf{20.37}
		& \textbf{39.46} & \underline{50.06} \\
		& CAPS (Top5-mean)
		& \underline{31.93} & \underline{19.54}
		& \underline{35.04} & \textbf{54.63} \\
		
		\bottomrule
	\end{tabular}
\end{table}

Table~\ref{tab:multi_results} shows strong performance from the
variable-wise extension despite its lack of explicit
cross-variable modeling.
CAPS (Max) achieves the best result on 11 of the 16
dataset--metric pairs, while CAPS (Top5-mean) achieves the best
result on four.
TranAD achieves the highest Affiliation-F on SMD.

Max aggregation performs better on most metrics for MSL, PSM,
and SMAP, whereas Top5-mean is stronger on three of the four
metrics for SMD.
These differences illustrate the effect of score aggregation:
Max emphasizes the strongest individual response, while
Top5-mean combines evidence from several variables.
Neither strategy is uniformly preferable across the evaluated
datasets and metrics.

The results support the usefulness of recovered temporal
supervision in a variable-wise multivariate pipeline.
Each detector learns from reference-anchored pairs for its
corresponding variable, and aggregation combines the resulting
anomaly evidence at the decision level.

This construction cannot explicitly model anomalies expressed
only through changing correlations or conditional relationships
among otherwise plausible individual trajectories.
The experiment therefore establishes a practical extension of
the temporal supervision-recovery procedure, while leaving
cross-variable anomaly structure unmodeled.
Incorporating such dependencies into representation learning
and counterpart construction remains a direction for future work.

\begin{sidewaystable}[p]
	\centering
	\small
	\setlength{\tabcolsep}{4pt}
	\renewcommand{\arraystretch}{1.15}
	\caption{Performance variability of CAPS and controlled TCN
		references across nine datasets. Per-dataset results are
		reported as mean $\pm$ standard deviation over five independent
		runs. Avg.\ is the unweighted average of the nine dataset means.
		Higher is better. The best mean among the methods listed in
		each dataset--metric group is shown in bold.}
	\label{tab:controlled_results_std}
	\resizebox{\linewidth}{!}{%
		\begin{tabular}{ll*{10}{c}}
			\toprule
			Metric & Method & IOPS & MGAB & NAB & Power & SED
			& UCR & NEK & TODS & YAHOO & Avg. \\
			\midrule
			
			\multirow{5}{*}{Affiliation-F}
			& TCN-AE
			& $83.91 \pm 0.80$ & $67.61 \pm 1.69$
			& $85.74 \pm 1.66$ & $83.34 \pm 0.91$
			& $67.75 \pm 1.08$ & $74.54 \pm 0.43$
			& $85.74 \pm 1.43$ & $72.61 \pm 0.42$
			& $82.47 \pm 0.43$ & 78.19 \\
			& TCN-Injection
			& $87.58 \pm 0.72$ & $69.13 \pm 1.56$
			& $88.89 \pm 1.37$ & $85.04 \pm 0.41$
			& $70.94 \pm 0.46$ & $84.32 \pm 1.40$
			& $81.52 \pm 0.84$ & $72.05 \pm 0.20$
			& $83.22 \pm 1.66$ & 80.30 \\
			& TCN-Supervised
			& $84.35 \pm 0.70$ & $67.59 \pm 0.98$
			& $89.62 \pm 0.97$ & $85.96 \pm 1.36$
			& $68.47 \pm 1.71$ & $72.40 \pm 1.12$
			& $85.24 \pm 1.40$ & $71.75 \pm 0.67$
			& $83.11 \pm 1.05$ & 78.72 \\
			& TCN-Forecast
			& $76.05 \pm 1.47$ & $67.27 \pm 0.95$
			& $84.06 \pm 1.09$ & $84.80 \pm 1.51$
			& $67.68 \pm 0.55$ & $72.81 \pm 0.92$
			& $85.83 \pm 0.81$ & $72.48 \pm 0.24$
			& $82.64 \pm 1.18$ & 77.07 \\
			& CAPS
			& $\mathbf{89.24 \pm 0.29}$ & $\mathbf{69.22 \pm 0.70}$
			& $\mathbf{93.25 \pm 0.62}$ & $\mathbf{85.98 \pm 0.97}$
			& $\mathbf{74.09 \pm 0.88}$ & $\mathbf{88.41 \pm 0.66}$
			& $\mathbf{86.30 \pm 0.28}$ & $\mathbf{86.85 \pm 0.97}$
			& $\mathbf{94.63 \pm 0.60}$ & \textbf{85.33} \\
			
			\midrule
			
			\multirow{5}{*}{$\mathrm{F1}_{\mathrm{T}}$}
			& TCN-AE
			& $39.38 \pm 1.02$ & $3.25 \pm 0.19$
			& $36.63 \pm 1.23$ & $19.92 \pm 0.52$
			& $10.23 \pm 0.28$ & $13.87 \pm 0.26$
			& $68.58 \pm 1.43$ & $24.85 \pm 0.21$
			& $58.94 \pm 0.97$ & 30.63 \\
			& TCN-Injection
			& $48.37 \pm 0.92$ & $\mathbf{23.89 \pm 0.84}$
			& $46.91 \pm 0.43$ & $27.04 \pm 0.35$
			& $21.59 \pm 0.76$ & $\mathbf{44.79 \pm 0.57}$
			& $64.45 \pm 1.00$ & $17.26 \pm 0.20$
			& $35.68 \pm 0.50$ & 36.66 \\
			& TCN-Supervised
			& $41.92 \pm 0.35$ & $1.94 \pm 0.17$
			& $42.50 \pm 0.39$ & $20.13 \pm 0.35$
			& $11.01 \pm 0.49$ & $9.56 \pm 0.55$
			& $72.93 \pm 0.80$ & $22.54 \pm 0.38$
			& $59.52 \pm 0.97$ & 31.34 \\
			& TCN-Forecast
			& $30.24 \pm 0.44$ & $1.56 \pm 0.05$
			& $37.01 \pm 0.68$ & $19.82 \pm 0.82$
			& $10.15 \pm 0.21$ & $8.50 \pm 0.28$
			& $68.89 \pm 0.75$ & $23.42 \pm 0.22$
			& $61.61 \pm 1.29$ & 29.02 \\
			& CAPS
			& $\mathbf{60.62 \pm 1.04}$ & $7.58 \pm 0.45$
			& $\mathbf{57.90 \pm 0.26}$ & $\mathbf{29.20 \pm 1.03}$
			& $\mathbf{24.27 \pm 0.31}$ & $41.53 \pm 0.22$
			& $\mathbf{80.15 \pm 0.08}$ & $\mathbf{46.17 \pm 0.78}$
			& $\mathbf{71.35 \pm 1.02}$ & \textbf{46.53} \\
			
			\midrule
			
			\multirow{5}{*}{Standard-F1}
			& TCN-AE
			& $27.15 \pm 0.60$ & $2.26 \pm 0.61$
			& $28.53 \pm 0.72$ & $20.08 \pm 0.53$
			& $10.22 \pm 0.39$ & $9.06 \pm 0.50$
			& $58.61 \pm 0.50$ & $20.72 \pm 0.18$
			& $49.27 \pm 1.09$ & 25.10 \\
			& TCN-Injection
			& $46.37 \pm 0.68$ & $\mathbf{14.61 \pm 0.66}$
			& $38.00 \pm 0.89$ & $27.00 \pm 0.40$
			& $21.42 \pm 0.74$ & $35.65 \pm 0.84$
			& $60.35 \pm 1.36$ & $16.60 \pm 0.30$
			& $31.21 \pm 0.50$ & 32.36 \\
			& TCN-Supervised
			& $42.38 \pm 0.93$ & $1.56 \pm 0.05$
			& $34.35 \pm 0.74$ & $20.03 \pm 0.51$
			& $10.92 \pm 0.33$ & $5.24 \pm 0.29$
			& $62.58 \pm 1.29$ & $19.23 \pm 0.43$
			& $51.24 \pm 1.26$ & 27.50 \\
			& TCN-Forecast
			& $18.65 \pm 0.53$ & $1.30 \pm 0.13$
			& $28.56 \pm 0.53$ & $19.77 \pm 0.67$
			& $10.16 \pm 0.36$ & $4.94 \pm 0.39$
			& $61.32 \pm 1.02$ & $20.56 \pm 0.26$
			& $54.63 \pm 1.31$ & 24.43 \\
			& CAPS
			& $\mathbf{51.63 \pm 0.59}$ & $4.54 \pm 0.45$
			& $\mathbf{51.78 \pm 0.22}$ & $\mathbf{29.37 \pm 1.71}$
			& $\mathbf{24.11 \pm 0.21}$ & $\mathbf{37.70 \pm 0.18}$
			& $\mathbf{76.23 \pm 0.56}$ & $\mathbf{41.24 \pm 0.45}$
			& $\mathbf{69.58 \pm 1.02}$ & \textbf{42.91} \\
			
			\midrule
			
			\multirow{5}{*}{VUS-PR}
			& TCN-AE
			& $17.27 \pm 0.59$ & $0.87 \pm 0.07$
			& $21.56 \pm 0.55$ & $14.95 \pm 0.23$
			& $8.99 \pm 0.53$ & $3.63 \pm 0.30$
			& $61.04 \pm 1.04$ & $46.30 \pm 0.23$
			& $41.03 \pm 0.60$ & 23.96 \\
			& TCN-Injection
			& $48.69 \pm 1.16$ & $\mathbf{3.27 \pm 0.18}$
			& $34.97 \pm 1.02$ & $\mathbf{21.87 \pm 0.68}$
			& $17.50 \pm 0.30$ & $26.07 \pm 0.86$
			& $60.76 \pm 0.77$ & $49.35 \pm 0.25$
			& $42.35 \pm 0.82$ & 33.87 \\
			& TCN-Supervised
			& $33.62 \pm 0.99$ & $0.61 \pm 0.06$
			& $30.66 \pm 0.82$ & $11.85 \pm 0.28$
			& $8.50 \pm 0.35$ & $2.87 \pm 0.16$
			& $67.52 \pm 0.98$ & $76.61 \pm 0.23$
			& $44.28 \pm 0.76$ & 30.72 \\
			& TCN-Forecast
			& $11.19 \pm 0.35$ & $0.57 \pm 0.05$
			& $20.85 \pm 0.48$ & $11.25 \pm 0.25$
			& $8.92 \pm 0.34$ & $2.37 \pm 0.15$
			& $63.35 \pm 0.62$ & $45.72 \pm 0.39$
			& $43.27 \pm 0.73$ & 23.05 \\
			& CAPS
			& $\mathbf{49.00 \pm 0.60}$ & $1.55 \pm 0.23$
			& $\mathbf{52.35 \pm 0.66}$ & $19.42 \pm 1.45$
			& $\mathbf{20.09 \pm 0.19}$ & $\mathbf{28.47 \pm 0.14}$
			& $\mathbf{80.52 \pm 0.76}$ & $\mathbf{86.26 \pm 1.17}$
			& $\mathbf{83.70 \pm 1.21}$ & \textbf{46.82} \\
			
			\bottomrule
		\end{tabular}%
	}
\end{sidewaystable}



\clearpage
\input{checklist.tex}

\end{document}

%% file: checklist.tex
\section*{NeurIPS Paper Checklist}

\begin{enumerate}
	
	\item {\bf Claims}
	\item[] Question: Do the main claims made in the abstract and introduction accurately reflect the paper's contributions and scope?
	\item[] Answer: \answerYes{}
	\item[] Justification: The abstract and introduction describe CAPS as a supervision-recovery framework based on transferable anomaly semantics, target-conditioned realization, and context-anchored detector supervision. The problem formulation and experiments clarify the assumptions, scope, and distinction between standard CAPS and the optional Diagnosis setting.
	
	\item {\bf Limitations}
	\item[] Question: Does the paper discuss the limitations of the work performed by the authors?
	\item[] Answer: \answerYes{}
	\item[] Justification: The conclusion and supplementary analyses discuss dependence on simulated semantic coverage, sensitivity to contaminated target references, and the lack of explicit cross-variable dependency modeling. Additional experiments examine these limitations and the computational cost of supervision recovery.
	\item[] Guidelines:
	\begin{itemize}
		\item The answer \answerNA{} means that the paper has no limitation while the answer \answerNo{} means that the paper has limitations, but those are not discussed in the paper.
		\item The authors are encouraged to create a separate ``Limitations'' section in their paper.
		\item The paper should point out any strong assumptions and how robust the results are to violations of these assumptions.
		\item The authors should reflect on the scope of the claims made.
		\item The authors should reflect on the factors that influence the performance of the approach.
		\item The authors should discuss the computational efficiency of the proposed algorithms and how they scale with dataset size.
		\item If applicable, the authors should discuss possible limitations of their approach to address problems of privacy and fairness.
		\item The authors should use their best judgment and recognize that individual actions in favor of transparency play an important role in developing norms that preserve the integrity of the community.
	\end{itemize}
	
	\item {\bf Theory assumptions and proofs}
	\item[] Question: For each theoretical result, does the paper provide the full set of assumptions and a complete (and correct) proof?
	\item[] Answer: \answerYes{}
	\item[] Justification: Section~\ref{sec:analysis} states Proposition~\ref{prop:transfer} with its finite-risk, integrability, and uniform Lipschitz assumptions. Appendix~A provides the proof, the derivation of the contextual realization gap, and additional discussion of the regularity conditions, learning-error term, and scope of the analysis.
	\item[] Guidelines:
	\begin{itemize}
		\item The answer \answerNA{} means that the paper does not include theoretical results.
		\item All the theorems, formulas, and proofs in the paper should be numbered and cross-referenced.
		\item All assumptions should be clearly stated or referenced in the statement of any theorems.
		\item The proofs can either appear in the main paper or the supplemental material.
		\item Any informal proof provided in the core of the paper should be complemented by formal proofs provided in appendix or supplemental material.
		\item Theorems and Lemmas that the proof relies upon should be properly referenced.
	\end{itemize}
	
	\item {\bf Experimental result reproducibility}
	\item[] Question: Does the paper fully disclose all the information needed to reproduce the main experimental results of the paper to the extent that it affects the main claims and/or conclusions of the paper (regardless of whether the code and data are provided or not)?
	\item[] Answer: \answerYes{}
	\item[] Justification: The method sections and Algorithm~\ref{alg:caps} describe representation learning, semantic-prior estimation, residual-expert training, counterpart construction, detector training, and inference. Appendices~\ref{appendix:implementation} and~\ref{appendix:exp_setup} provide implementation and evaluation details, complemented by the released code and experiment configurations.
	\item[] Guidelines:
	\begin{itemize}
		\item The answer \answerNA{} means that the paper does not include experiments.
		\item If the paper includes experiments, a \answerNo{} answer to this question will not be perceived well by the reviewers.
		\item If the contribution is a dataset and\slash or model, the authors should describe the steps taken to make their results reproducible or verifiable.
		\item Reproducibility can be accomplished in various ways, including code release, detailed implementation instructions, model checkpoints, or other appropriate means.
		\item If the contribution is primarily a new algorithm, the paper should make it clear how to reproduce that algorithm.
		\item If the contribution is primarily a new model architecture, the paper should describe the architecture clearly and fully.
		\item If the contribution is a new model, there should either be a way to access this model for reproducing the results or a way to reproduce the model.
	\end{itemize}
	
	\item {\bf Open access to data and code}
	\item[] Question: Does the paper provide open access to the data and code, with sufficient instructions to faithfully reproduce the main experimental results, as described in supplemental material?
	\item[] Answer: \answerYes{}
	\item[] Justification: Appendix~\ref{appendix:implementation} links the repository containing the CAPS implementation, training and evaluation scripts, experiment configurations, and data-preparation instructions. The released simulated normal--anomalous pairs include mode assignments, anomaly supports, and reference identifiers.
	\item[] Guidelines:
	\begin{itemize}
		\item The answer \answerNA{} means that paper does not include experiments requiring code.
		\item Please see the NeurIPS code and data submission guidelines for more details.
		\item While we encourage the release of code and data, we understand that this might not be possible, so \answerNo{} is an acceptable answer.
		\item The instructions should contain the exact command and environment needed to run to reproduce the results.
		\item The authors should provide instructions on data access and preparation.
		\item The authors should provide scripts to reproduce all experimental results for the new proposed method and baselines.
		\item At submission time, to preserve anonymity, the authors should release anonymized versions if applicable.
	\end{itemize}
	
	\item {\bf Experimental setting/details}
	\item[] Question: Does the paper specify all the training and test details (e.g., data splits, hyperparameters, how they were chosen, type of optimizer) necessary to understand the results?
	\item[] Answer: \answerYes{}
	\item[] Justification: Section~5 and Appendices~\ref{appendix:implementation} and~\ref{appendix:exp_setup} specify the data splits, preprocessing, metrics, threshold selection, architecture, and optimization settings. Mode organization and Diagnosis are described in Appendix~\ref{appendix:mode_details}, with additional comparison protocols provided in the corresponding experimental subsections.
	
	\item {\bf Experiment statistical significance}
	\item[] Question: Does the paper report error bars suitably and correctly defined or other appropriate information about the statistical significance of the experiments?
	\item[] Answer: \answerYes{}
	\item[] Justification: The appendix reports per-dataset means and standard deviations over five independent runs for CAPS and the controlled TCN references. These statistics quantify variability across repeated runs. The reported uncertainty is the standard deviation, and the comparisons are not presented as formal significance tests.
	\item[] Guidelines:
	\begin{itemize}
		\item The answer \answerNA{} means that the paper does not include experiments.
		\item The authors should answer \answerYes{} if the results are accompanied by error bars, confidence intervals, or statistical significance tests, at least for the experiments that support the main claims of the paper.
		\item The factors of variability that the error bars are capturing should be clearly stated.
		\item The method for calculating the error bars should be explained.
		\item The assumptions made should be given.
		\item It should be clear whether the error bar is the standard deviation or the standard error of the mean.
		\item If error bars are reported in tables or plots, the authors should explain in the text how they were calculated and reference the corresponding figures or tables in the text.
	\end{itemize}
	
	\item {\bf Experiments compute resources}
	\item[] Question: For each experiment, does the paper provide sufficient information on the computer resources (type of compute workers, memory, time of execution) needed to reproduce the experiments?
	\item[] Answer: \answerYes{}
	\item[] Justification: Implementation settings and computational costs are provided in Appendices~\ref{appendix:implementation} and~\ref{appendix:efficiency}, covering simulated-domain training, counterpart generation, and detector inference.
	\item[] Guidelines:
	\begin{itemize}
		\item The answer \answerNA{} means that the paper does not include experiments.
		\item The paper should indicate the type of compute workers CPU or GPU, internal cluster, or cloud provider, including relevant memory and storage.
		\item The paper should provide the amount of compute required for each of the individual experimental runs as well as estimate the total compute.
		\item The paper should disclose whether the full research project required more compute than the experiments reported in the paper.
	\end{itemize}
	
	\item {\bf Code of ethics}
	\item[] Question: Does the research conducted in the paper conform, in every respect, with the NeurIPS Code of Ethics \url{https://neurips.cc/public/EthicsGuidelines}?
	\item[] Answer: \answerYes{}
	\item[] Justification: The research uses public time-series anomaly detection benchmarks and simulated normal--anomalous pairs. It does not involve human-subject experiments or private personal-data collection, and is intended for methodological evaluation under the stated benchmark protocols.
	\item[] Guidelines:
	\begin{itemize}
		\item The answer \answerNA{} means that the authors have not reviewed the NeurIPS Code of Ethics.
		\item If the authors answer \answerNo{}, they should explain the special circumstances that require a deviation from the Code of Ethics.
		\item The authors should make sure to preserve anonymity.
	\end{itemize}
	
	\item {\bf Broader impacts}
	\item[] Question: Does the paper discuss both potential positive societal impacts and negative societal impacts of the work performed?
	\item[] Answer: \answerYes{}
	\item[] Justification: The paper discusses potential benefits for monitoring applications and potential risks associated with false alarms and missed anomalies.
	\item[] Guidelines:
	\begin{itemize}
		\item The answer \answerNA{} means that there is no societal impact of the work performed.
		\item If the authors answer \answerNA{} or \answerNo{}, they should explain why their work has no societal impact or why the paper does not address societal impact.
		\item Examples of negative societal impacts include potential malicious or unintended uses, fairness considerations, privacy considerations, and security considerations.
		\item The conference expects that many papers will be foundational research and not tied to particular applications, let alone deployments.
		\item The authors should consider possible harms that could arise when the technology is used as intended and functioning correctly, when it gives incorrect results, and through misuse.
		\item If there are negative societal impacts, the authors could also discuss possible mitigation strategies.
	\end{itemize}
	
\item {\bf Safeguards}
\item[] Question: Does the paper describe safeguards that have been put in place for responsible release of data or models that have a high risk for misuse (e.g., pre-trained language models, image generators, or scraped datasets)?
\item[] Answer: \answerNA{}
\item[] Justification: The released assets concern time-series anomaly detection code and simulated anomaly pairs. The work does not introduce high-risk general-purpose models or datasets containing scraped personal information.
\item[] Guidelines:
\begin{itemize}
	\item The answer \answerNA{} means that the paper poses no such risks.
	\item Released models that have a high risk for misuse or dual-use should be released with necessary safeguards to allow for controlled use of the model.
	\item Datasets that have been scraped from the Internet could pose safety risks.
	\item We recognize that providing effective safeguards is challenging, and many papers do not require this.
\end{itemize}
	\item {\bf Licenses for existing assets}
	\item[] Question: Are the creators or original owners of assets (e.g., code, data, models), used in the paper, properly credited and are the license and terms of use explicitly mentioned and properly respected?
	\item[] Answer: \answerYes{}
	\item[] Justification: The datasets and baseline implementations are credited to their original sources and used in accordance with their respective licenses and terms of use.
	\item[] Guidelines:
	\begin{itemize}
		\item The answer \answerNA{} means that the paper does not use existing assets.
		\item The authors should cite the original paper that produced the code package or dataset.
		\item The authors should state which version of the asset is used and, if possible, include a URL.
		\item The name of the license should be included for each asset.
		\item For scraped data from a particular source, the copyright and terms of service of that source should be provided.
		\item If assets are released, the license, copyright information, and terms of use in the package should be provided.
		\item For existing datasets that are re-packaged, both the original license and the license of the derived asset should be provided.
		\item If this information is not available online, the authors are encouraged to reach out to the asset's creators.
	\end{itemize}
	
	\item {\bf New assets}
	\item[] Question: Are new assets introduced in the paper well documented and is the documentation provided alongside the assets?
	\item[] Answer: \answerYes{}
	\item[] Justification: The repository linked in Appendix~\ref{appendix:implementation} documents the CAPS implementation and released simulated pairs, including their mode assignments, anomaly supports, and reference identifiers. Training, evaluation, and data-preparation instructions accompany these assets.
	\item[] Guidelines:
	\begin{itemize}
		\item The answer \answerNA{} means that the paper does not release new assets.
		\item Researchers should communicate the details of the dataset\slash code\slash model as part of their submissions via structured templates.
		\item The paper should discuss whether and how consent was obtained from people whose asset is used.
		\item At submission time, remember to anonymize your assets if applicable.
	\end{itemize}
	
	\item {\bf Crowdsourcing and research with human subjects}
	\item[] Question: For crowdsourcing experiments and research with human subjects, does the paper include the full text of instructions given to participants and screenshots, if applicable, as well as details about compensation (if any)?
	\item[] Answer: \answerNA{}
	\item[] Justification: The paper does not involve crowdsourcing, user studies, or research with human subjects.
	\item[] Guidelines:
	\begin{itemize}
		\item The answer \answerNA{} means that the paper does not involve crowdsourcing nor research with human subjects.
		\item Including this information in the supplemental material is fine.
		\item According to the NeurIPS Code of Ethics, workers involved in data collection, curation, or other labor should be paid at least the minimum wage in the country of the data collector.
	\end{itemize}
	
	\item {\bf Institutional review board (IRB) approvals or equivalent for research with human subjects}
	\item[] Question: Does the paper describe potential risks incurred by study participants, whether such risks were disclosed to the subjects, and whether Institutional Review Board (IRB) approvals (or an equivalent approval/review based on the requirements of your country or institution) were obtained?
	\item[] Answer: \answerNA{}
	\item[] Justification: The paper does not involve crowdsourcing or research with human subjects, so institutional human-subject review is not applicable.
	\item[] Guidelines:
	\begin{itemize}
		\item The answer \answerNA{} means that the paper does not involve crowdsourcing nor research with human subjects.
		\item Depending on the country in which research is conducted, IRB approval or equivalent may be required for human subjects research.
		\item We expect authors to adhere to the NeurIPS Code of Ethics and the guidelines for their institution.
		\item For initial submissions, do not include any information that would break anonymity.
	\end{itemize}
	
	\item {\bf Declaration of LLM usage}
	\item[] Question: Does the paper describe the usage of LLMs if it is an important, original, or non-standard component of the core methods in this research? Note that if the LLM is used only for writing, editing, or formatting purposes and does \emph{not} impact the core methodology, scientific rigor, or originality of the research, declaration is not required.
	\item[] Answer: \answerNA{}
	\item[] Justification: LLMs are not an important, original, or non-standard component of the proposed method. Language-tool assistance is limited to writing, editing, or formatting and does not determine the method, experimental results, or theoretical analysis.
	\item[] Guidelines:
	\begin{itemize}
		\item The answer \answerNA{} means that the core method development in this research does not involve LLMs as any important, original, or non-standard components.
		\item Please refer to the LLM policy in the NeurIPS handbook for what should or should not be described.
	\end{itemize}
	
\end{enumerate}